\documentclass[journal]{IEEEtran}
\usepackage{array}
\newcolumntype{C}[1]{>{\centering}p{#1}}
\usepackage{amssymb}
\usepackage{amsmath}
\usepackage{amsthm}
\usepackage{algorithm}
\usepackage{algorithmic}
\usepackage{amsfonts} 
\usepackage{booktabs}
\usepackage{cite}
\usepackage{float} 
\usepackage[pdftex]{graphicx}
 \graphicspath{pics/}
 \DeclareGraphicsExtensions{.pdf,.jpeg,.png,.jpg}
\usepackage{color}
\usepackage[dvipsnames, svgnames, x11names]{xcolor}
\usepackage{lipsum}
\usepackage{multirow} 
\usepackage{pifont}
\usepackage{soul}
\usepackage{subfigure}  
\usepackage{times}
\usepackage{url}
\usepackage{verbatim}
\usepackage{fancyhdr} 

\begin{document}
\title{Unsupervised Deep Learning for IoT Time Series}

\author{Ya~Liu, Yingjie~Zhou, Kai~Yang, and Xin Wang
\thanks{Ya Liu and Kai Yang are with the Department of Computer Science and Technology, Tongji University, Shanghai, China (e-mail: yaliu@tongji.edu.cn, kaiyang@tongji.edu.cn). Yingjie Zhou is with College of Computer Science, Sichuan university, Chengdu, Sichuan, China (e-mail: yjzhou09@gmail.com; yjzhou@scu.edu.cn). Xin Wang is with School of Information Science and Engineering, Fudan University, Shanghai, China (e-mail: xwang11@fudan.edu.cn).}
}

\maketitle

\begin{abstract}
IoT time series analysis has found numerous applications in a wide variety of areas, ranging from health informatics to network security. Nevertheless, the complex spatial temporal dynamics and high dimensionality of IoT time series make the analysis increasingly challenging. In recent years, the powerful feature extraction and representation learning capabilities of deep learning (DL) have provided an effective means for IoT time series analysis. However, few existing surveys on time series have systematically discussed unsupervised DL-based methods. To fill this void, we investigate unsupervised deep learning for IoT time series, i.e., unsupervised anomaly detection and clustering, under a unified framework. We also discuss the application scenarios, public datasets, existing challenges, and future research directions in this area.
\end{abstract}


\IEEEpeerreviewmaketitle

\section{Introduction}
\markboth{The manuscript has been accepted by IEEE Internet of Things Journal. DOI: 10.1109/JIOT.2023.3243391}{The manuscript has been accepted by IEEE Internet of Things Journal. DOI: 10.1109/JIOT.2023.3243391}
\label{gen_inst}
\IEEEPARstart{W}{ith} the development of the fifth generation (5G) network, the Internet of Things (IoT) has become ubiquitous in our daily life. 5G enables connections with extraordinary speed, expanded bandwidth, and low latency, serving billions of mobile users and IoT devices \cite{yang2009auction}. It is estimated that the global economic impact of IoT will reach \$11.1 trillion per year by 2025 \cite{manyika2015internet}. Nowadays, IoT sensors continue to generate large amounts of time series data, which contain meaningful knowledge of the monitored system. Analyzing these time series data can help operators understand the underlying causes of systemic patterns over time and provide a better user experience with lower operating costs. Thus, time series analysis, e.g., anomaly detection and clustering, has been in great demand in many fields, from energy and finance to healthcare and IT Operations.

Traditional time series analysis methods have achieved favorable performance with hand-crafted features and sufficient expert knowledge. However, compared with non-IoT time series, IoT time series exhibit some unique characteristics which render traditional time series analysis methods not directly applicable. First, IoT time series can be of massive amount and high-dimensional as 5G and beyond communication systems allow monitoring of hundreds or even millions of IoT devices simultaneously, which entails scalability as a key challenge for IoT time series analysis \cite{stankovic2014research}. In addition, as opposed to non-IoT time series data with only temporal correlations, IoT time series exhibit not only temporal correlations but also complicated spatial correlations. That is because IoT devices are usually geographically close to each other \cite{mary2017imputing}. For example, in smart transportation, multiple sensors installed on vehicles are exploited to record vehicle's real-time information such as speed and position\cite{wang2018distributed}. Then control center plans convenient routes for users based on sensor time series to avoid traffic congestion.

Second, the spatial temporal dynamics of IoT time series can be extremely complex. IoT time series may exhibit various patterns over different spatial and temporal scales. Here are some examples: 1) IoT traffic time series triggered by events of programmed machine activities or human interventions exhibit both long-term and short-term temporal dependencies \cite{tahaei2020rise}. Specifically, frequent programmed machine activities, such as periodic updates, are almost periodic and constitute short-term dependencies. In contrast, human interventions, such as viewing surveillance videos, are bursty and occur much less frequently. In this scenario, IoT dynamics are unpredictable because they are intertwined with or even partially determined by human behavior \cite{duan2017human}. 2) Some IoT time series data may be non-stationary due to the influence of the complex environment \cite{zhang2019adaptive}, such as concept drift \cite{gama2014survey} and seasonality \cite{wen2019robuststl}. Intuitively, non-stationarity means that the statistical properties of the process generating the IoT time series change over time. 3) The sources of IoT time series data may be heterogeneous in the form of protocols, device data format, communication capabilities of the devices, technologies, and hardware\cite{vargas2016smart}. The heterogeneity of data sources can further lead to the heterogeneity of data characteristics, that is, time series generated by different IoT devices/services may exhibit different behaviors. 4) The low-cost, resource-constrained IoT sensors and the relatively uncontrollable environments in which they are deployed lead to more noise in IoT time series than data collected from typical hosts. Noise included in IoT time series may be caused by minor variations in the sensitivity of the detector, unrelated events occurring within the vicinity of the sensor, or transmission-based errors in the data management system \cite{cook2019anomaly}.

Nowadays, deep learning (DL) has been considered effective in the time series analysis \cite{Javaid2016ADL,Peng2015MultiScaleCI,Ma2019LearningRF}. The development of DL has enabled researchers to solve complex problems in an end-to-end fashion to avoid manual feature extraction \cite{gamboa2017deep}. In general, DL methods are categorized into supervised, semi-supervised, and unsupervised methods based on the labels available in the dataset. Since labeling large amounts of data requires human resources that most organizations cannot afford, unsupervised DL methods have been used in a wide range of applications in IoT scenarios \cite{bengio2012unsupervised, nomm2018unsupervised}.

\subsection{Existing Surveys}
\IEEEpubidadjcol  
Several researchers have conducted surveys on time series modeling and mining. This section summarizes the existing surveys in the literature and compares them with our work, as Table \ref{table1} shows. To the best of our knowledge, most of the existing surveys on time series analysis do not focus specifically on IoT systems. Furthermore, existing surveys have not systematically summarized the application of unsupervised DL in time series. In contrast, they only investigate specific machine learning (ML) tasks, such as anomaly detection \cite{choi2021deep, Gupta2014OutlierDF, islam2017time, cook2019anomaly, Braei2020AnomalyDI, BlazquezGarcia2020ARO}, classification\cite{susto2018time, ismail2019deep}, clustering \cite{Liao2005ClusteringOT, FrhwirthSchnatter2011PanelDA, Zolhavarieh2014ARO, Aghabozorgi2015TimeseriesC, Javed2020ABS, alqahtani2021deep} and prediction \cite{han2019review, lim2021time}. This article focuses on IoT time series and systematically discusses the advantages and applications of unsupervised DL methods. The most significant difference between this article and existing surveys is that we bring unsupervised anomaly detection and clustering into a unified perspective and provide a general unsupervised DL-based time series analysis framework for IoT. Studying unsupervised anomaly detection and clustering under this unified framework helps to make works in these two fields learn from each other, reveal the relationship between DL's capabilities and structures, and then improve the ability of DL methods to analyze IoT time series. 

\newcommand\cmark{\textcolor{Blue}{\ding{51}}} 
\begin{table*}[!t]
\renewcommand{\arraystretch}{1.2} 
\caption{summary of the related surveys}
\label{table1}
\centering
\begin{tabular}{c|c|p{0.57\textwidth}<{\centering}|c|c|p{0.07\textwidth}<{\centering}|c}
\hline \hline
\multirow{3}{*}{Year} & \multirow{3}{*}{Ref.} & \multirow{3}{*}{Contribution} & \multicolumn{4}{c}{Scope} \\
\cline{4-7}   
&&&\multirow{2}{*}{IoT} &\multirow{2}{*}{DL} & Anomaly Detection &\multirow{2}{*}{Clustering}\\
\hline
2014 & \cite{Gupta2014OutlierDF} & Overview of outlier detection techniques for various forms of temporal data. &&&\cmark & \\ 
\hline
2017 & \cite{islam2017time} & Review of graph theory-based anomaly detection in time series social networks data. & & & \cmark & \\
\hline
\multirow{2}{*}{2019} & \multirow{2}{*}{\cite{cook2019anomaly}} & Survey  of  current methods and future challenges of applying anomaly detection techniques to IoT data. & \multirow{2}{*}{\cmark} & \multirow{2}{*}{\cmark} & \multirow{2}{*}{\cmark} & \\
\hline
2020 & \cite{Braei2020AnomalyDI} & Research of statistical, ML and DL methods for univariate time series anomaly detection. & &\cmark & \cmark & \\
\hline
2021 & \cite{choi2021deep} & Review of DL-based anomaly detection methods for time series data. & & \cmark & \cmark & \\
\hline
2021 & \cite{BlazquezGarcia2020ARO} & Review of unsupervised outlier detection techniques in the context of time series. & & \cmark & \cmark & \\
\hline
2005 & \cite{Liao2005ClusteringOT} & Survey of  the algorithms, criteria and  applications of time series clustering. &&&& \cmark \\
\hline
2011 & \cite{FrhwirthSchnatter2011PanelDA} & Review of panel time series data clustering based on finite mixture models. &&&& \cmark \\
\hline
2014 & \cite{Zolhavarieh2014ARO} & Survey of various subsequence time series clustering approaches. &&&& \cmark \\
\hline
2015 & \cite{Aghabozorgi2015TimeseriesC} & Exposition of four main components of time series clustering. &&&& \cmark \\
\hline
2019 & \cite{Ali2019ClusteringAC} & Review of clustering or classification used in visual analytics for time series data. &&\cmark && \cmark\\
\hline
2020 & \cite{Javed2020ABS} & Research on the benchmark  of time series clustering. && \cmark && \cmark\\
\hline
\multirow{2}{*}{2021} & \multirow{2}{*}{\cite{alqahtani2021deep}} & Review of deep time series clustering (DTSC) with a case study in the context of movement behavior clustering. && \multirow{2}{*}{\cmark} && \multirow{2}{*}{\cmark} \\
\hline
\multicolumn{2}{c|}{\centering \multirow{2}{*}{Ours}} & Survey of unsupervised deep learning methods for IoT time series analysis. Unsupervised anomaly detection and clustering are investigated under a unified framework. & \multirow{2}{*}{\cmark} & \multirow{2}{*}{\cmark} & \multirow{2}{*}{\cmark} & \multirow{2}{*}{\cmark}\\
\hline
\end{tabular}
\end{table*}

\begin{figure}[!t]
\centering
\includegraphics[width=3.5in]{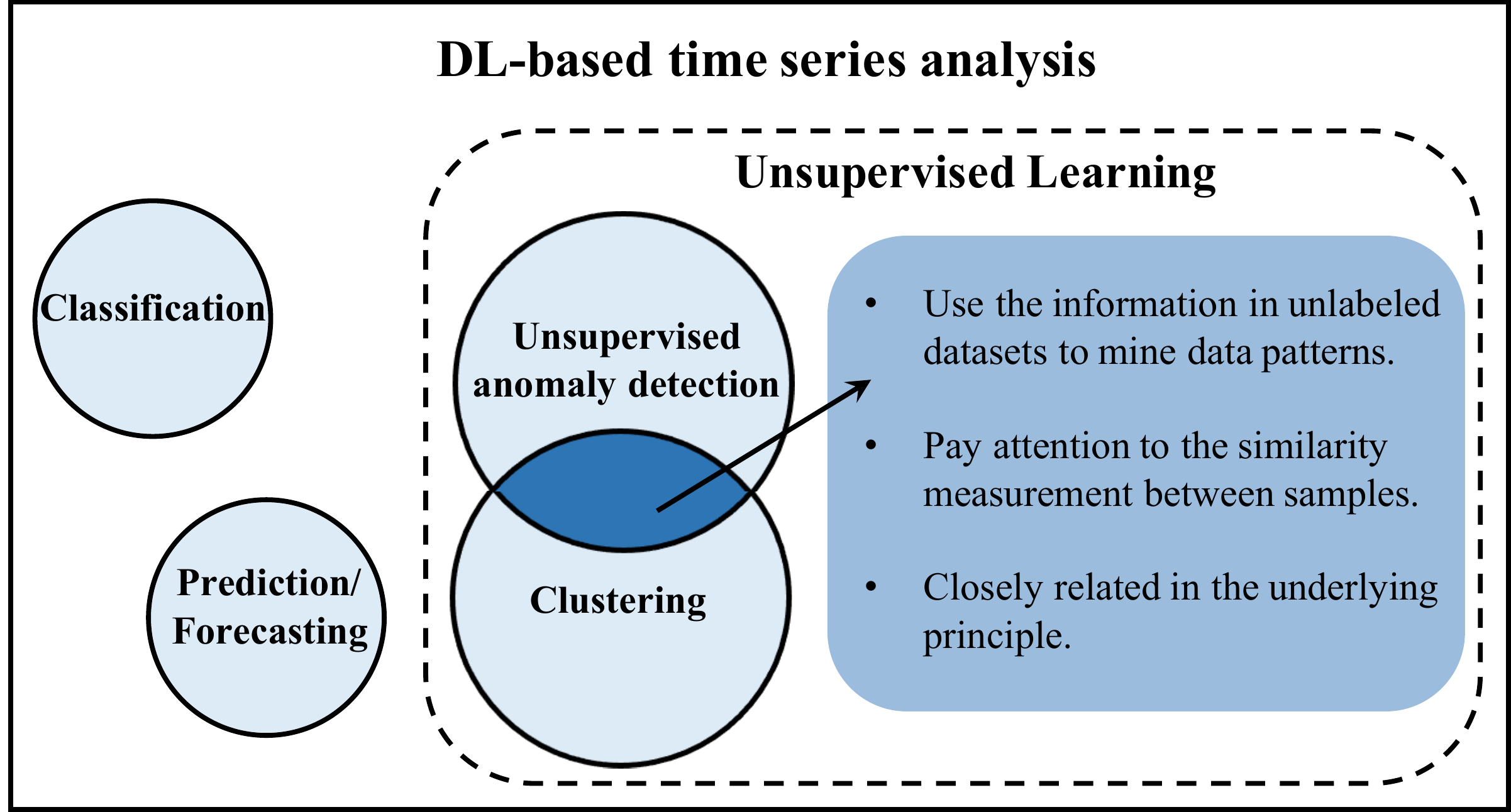}
\caption{The relevance between unsupervised anomaly detection and clustering.}
\label{fig1.0}
\end{figure}

This paper focuses on unsupervised deep learning for time series analysis with emphasis on anomaly detection and clustering. Other unsupervised time series modeling approaches based on statistical methods other than DL methods, such as hidden Markov model \cite{eddy1996hidden} and functional principal component analysis \cite{shang2014survey}, are less relevant to our topic and not included in the article. Our unique survey perspective is also supported by the relevance of unsupervised anomaly detection and clustering in the following aspects, as shown in Fig. \ref{fig1.0}. First, resulting from the prohibitive cost for accessing ground-truth labels of anomalies, anomaly detection methods in practice are predominately carried out in an unsupervised manner \cite{ruff2021unifying}. Clustering is also a typical unsupervised method \cite{ezugwu2022comprehensive}. That is to say, they both have to mine data patterns without supervision, so it is particularly important to make full use of the information in the data. Second, unsupervised anomaly detection and clustering both rely on the similarity measurement of samples when mining data patterns. Specifically, anomaly detection methods identify anomalies by measuring the similarity between the features of unknown samples and those of normal samples. The clustering, on the other hand, aims to organize samples with similar features into the same group. The above analysis shows that unsupervised anomaly detection and clustering are closely related in the underlying principle. In fact, there have been some works using clustering to find anomalies in unlabeled data \cite{wang2011two, pu2020hybrid}. Typically, they cluster the data samples first, and then assign anomaly score by using the distance between the samples and the cluster centers \cite{aggarwal2017introduction}.

This article investigates unsupervised anomaly detection and clustering in a unifying view to provide a general framework for unsupervised DL-based time series analysis in the context of IoT. We then organize and discuss current works along this framework, focusing on the structures and capabilities of DL models. In addition, we also discuss emerging application scenarios, public datasets, challenges, and potential directions for IoT time series analysis to enhance the breadth of this survey.

\subsection{Scope and Organization}
The scope of this review is as follows: First we discuss the motivation for using DL techniques in light of the requirements of IoT time series analysis. Then we investigate two different tasks, anomaly detection and clustering, in a unified manner to summarize the general flow of analyzing time series using DL techniques. After that, we introduce the current state-of-the-art DL techniques and discuss their role in each stage of the time series analysis flow. In addition, we also list emerging applications and public datasets for IoT time series. Finally, we discuss existing research challenges and future directions. To the best of our knowledge, this paper is the first survey of unsupervised DL methods for IoT time series.

\emph{Article Organization:} The remainder of this article is organized as follows. Section \ref{background} first introduces the concept of time series analysis, which mainly focuses on time series anomaly detection and clustering. After that, we discuss the challenges of IoT time series analysis and the motivation of using DL. Section \ref{UniModel} introduces a unified framework of unsupervised DL-based IoT time series analysis, which mainly consists of three subsections. The first subsection is about data pre-processing of IoT time series data. The rest parts are feature extraction and pattern identification based on DL methods.

A detailed review of DL methods related to time series analysis is provided in Section \ref{deepmodels}. General purpose DL models and techniques are surveyed first. Then state-of-the-art DL-based methods for time series feature extraction, anomaly detection, and clustering are classified and surveyed in detail. Section \ref{Applications} summarizes the applications and datasets of IoT time series analysis. Section \ref{Directions} discusses challenges and future research directions. Finally, we summarize our work in Section \ref{Conclusion}.

\section{Motivation for Using DL in IoT Time Series Analysis}
\label{background}
This section discusses the motivation for using DL in IoT time series analysis. We first introduce time series analysis, focusing mainly on anomaly detection and clustering. Then we touch upon the challenges of analyzing IoT time series. Finally, we establish the motivation for using DL.

\subsection{Time Series Analysis}
A time series is a set of observations $\boldsymbol{x}_{t}$, each one being recorded at a specified time instance $t$ \cite{brockwell2009time}. Compared with other types of data, time series data contain complex temporal dependencies and are often high-dimensional, making them challenging to model and analyze \cite{langkvist2014review}. Time series analysis mainly includes classification, forecasting/prediction, anomaly detection, and clustering. We focus on unsupervised time series anomaly detection and clustering in this article.

\subsubsection{Time Series Anomaly Detection}
An anomaly is defined as an observation that deviates significantly from the majority of data \cite{Hawkins1980IdentificationOO}. It usually has actionable pieces of information which could be meaningful \cite{al2021review}. The basic interpretation of anomaly detection is to identify patterns that do not conform to the expected behaviors of the system \cite{Chandola2009AnomalyDA}. In the context of IoT, a general definition of an anomaly is ``a measurable consequence of an unexpected change in the state of a system that is outside of its local or global norm'' \cite{cook2019anomaly}. Anomalies in the IoT system may come from cyber-attacks, system failures, noise, etc. For example, a sudden increase in the temperature of a room in a factory can signal that there is a fire. Anomaly detection, therefore, is the first step to secure IoT systems and has become an important research area \cite{xie2016distributed, armonchange, active-IoT-IDS}.

Following the literature, there are two ways to categorize anomalies in time series. First, time series anomalies can be divided into point anomalies, subsequence anomalies, and sequence anomalies depending on the granularity \cite{Gupta2014OutlierDF, BlazquezGarcia2020ARO}.
\begin{itemize} 
\item Point Anomalies. Point anomalies are data points that show significant deviations from other points in the time series (global point anomalies) or from their neighboring points in a particular frame (local point anomalies). Such point anomalies may be caused by noise, sensor failures, or short-term outages in the system.

\item Subsequence Anomalies. A subsequence is a set of consecutive observations within a time series. Subsequence anomalies are subsequences that deviate from the expected patterns. However, if viewed separately, individual points of the subsequences may all be within the expected range. 

\item Sequence Anomalies. When the input data is multivariable, a univariate time series whose behavior is significantly different from others is deemed as a sequence anomaly.
\end{itemize}

Second, from a behavior perspective, time series anomalies can be divided into point anomalies, contextual anomalies, and collective anomalies \cite{choi2021deep, cook2019anomaly}. 
\begin{itemize}
\item Point Anomalies. Here, point anomalies refer to observations or sequences that abruptly deviate from the normal state of the entire dataset.
\item Contextual Anomalies. Contextual anomalies are observations or sequences which not deviate from the normal range in a global perspective but are out of the expected pattern when considering the given context. 
\item Collective Anomalies. Sets of observations showing distinct patterns relative to the rest of the data are considered collective anomalies. 
\end{itemize}

Anomaly detection is typically categorized into three aspects according to the input type: supervised, unsupervised, and semi-supervised. Among them, the essence of a supervised anomaly detection problem is a classification problem that distinguishes abnormality from normality. In practice, the scarcity of abnormal events limits supervised methods. Semi-supervised anomaly detection learns a model of the normal class and anomalies can be detected afterwards by deviating from that model \cite{goldstein2016comparative}. Unsupervised anomaly detection gains no access to labels and identifies the shared patterns among the data instances to uncover the anomalies.

\subsubsection{Time Series Clustering}
Clustering is one of the most commonly used unsupervised learning algorithms. Time series clustering has been widely used in economics, medicine, engineering, and other fields. The goal of clustering is to organize objects into homogeneous groups where the intra-group similarities are maximized, and the inter-group similarities are minimized \cite{chen1996data}. Traditional clustering methods are generally divided into five categories \cite{han2011data}: partitioning, hierarchical, density-based, grid-based, and model-based methods. 

However, due to the characteristics of time series data, it is difficult for traditional clustering methods to achieve good performance on raw time series. There are two main strategies to adapt traditional clustering methods to time series data \cite{Liao2005ClusteringOT}. 

The first strategy is to choose a specific distance metric for time series data, such as Euclidean distance, Mahalanobis distance, dynamic time warping (DTW) distance, and Kullback–Liebler distance \cite{Liao2005ClusteringOT}. Among these methods, Euclidean distance is the most widely used metric. But it is not suitable for multivariable IoT time series due to its ill-defined concept of proximity on high-dimensional settings \cite{maldonado2019alternative}. Kullback–Liebler distance can effectively describe the similarity between different distributions, and it regards time series as probability distributions. However, Euclidean distance and Kullback–Liebler distance require that the lengths of all time series must be equal, which is not applicable in many real IoT situations. Mahalanobis distance is a measure of the distance between a variable and a distribution which is calculated by a mean and the covariance matrix. This metric has advantages in modeling multivariate time series as it takes into account the correlations of different variables \cite{mei2015learning}. Besides, Mahalanobis distance is robust to missing values \cite{sitaram2015measure}, which are common in IoT time series. DTW is a mapping of points between a pair of time series designed to minimize the pairwise Euclidean distance. DTW tries to warp the time of the two time series to find the closest possible match \cite{kang2020fpga}. Therefore, DTW is effective at finding similar time series with time shifts. Such shifts can be observed in IoT time series data due to physical misplacement or other erroneous acts.

The second strategy is to extract features from time series and subsequently cluster the time series based on the extracted features. This article mainly focuses on the second strategy in the context of DL.

\subsection{Challenges of IoT Time Series Analysis}
The term Internet of Things (IoT) generally refers to scenarios where network connectivity and computing capability extend to sensors and everyday items, allowing these devices to generate and exchange data with minimal human intervention \cite{rose2015internet}. Examples of IoT machines and systems could be manufacturing, wearable devices, or smart cities. These IoT devices continuously generate a large number of multi-dimensional time series and store critical information \cite{al2020survey}. Examining these collected data is of great significance for system security and resource optimization. For example, detecting suspicious events from time series can reduce threats and avoid unseen issues that cause downtime in the applications, allowing administrators to minimize losses \cite{al2021review}. Besides, clustering daily electricity time series can mine the correlation information between different buildings in an area, which provides a basis for optimizing the electricity price setting and power facility configuration \cite{li2019electricity}.

However, the complicated spatial temporal dynamics, high dimensionality, and large volume of IoT time series inevitably entail challenges for data analysis based on traditional machine learning methods. First, traditional machine learning algorithms fail to capture the spatial and temporal correlations simultaneously \cite{dai2020spatio}. Most traditional algorithms only focus on the temporal correlations of IoT time series, with no or limited addressing of the spatial impact among IoT devices. Second, scalability is a key challenge for IoT time series analysis. Dealing with the complex spatial temporal dynamics of high-dimensional, large-amount IoT time series exceeds the capabilities of traditional methods, which rely on expensive, time consuming-manual feature extraction and prior expert knowledge. In addition to computational efficiency, scalability is also about lower communication overhead (e.g., how often a device needs to communicate with other machines), as well as reduced information needed (e.g., what type of information a device needs before making decisions) \cite{chen2019learning}.

\subsection{Deep Learning for IoT Time Series Analysis}
Machine Learning (ML) is intended to allow a system to learn from the past or the present and to use the knowledge to make future predictions or decisions \cite{habeeb2019real}. Deep learning (DL) is a subfield of ML which enables computational models composed of multiple processing layers to learn data representations. The multiple levels of features in DL are automatically discovered and composed together to produce the outputs. Compared with ML, the main advantage of DL is the automatic feature extraction that avoids the tedious labor of generating feature representations manually. DL has gained great recognition in many areas such as computer vision, natural language processing, and bioinformatics. Nowadays, academia and industry apply DL to wider applications, such as IoT scenarios. IoT networks produce a large amount of data, and therefore, traditional data collection, storage, and processing techniques may not work at this scale \cite{hussain2020machine}. However, these data are required by DL approaches to bring intelligence to the systems. 

Applying DL methods to IoT time series analysis has the following advantages. First, DL methods achieve higher power and flexibility when dealing with massive and high-dimensional IoT time series due to their ability to process large amounts of data in parallel. Second, DL has powerful automatic feature extraction capabilities. Theoretically, DL models can approximate any complex non-linear functions and can fit any curves as long as they have enough layers and neurons \cite{wang2020deep}. With the multi-layer structure of neural networks, the complex spatial temporal dynamics in IoT time series data can be learned automatically and effectively.

However, using DL techniques in IoT time series analysis brings additional challenges. First, it is challenging to label large amounts of data. Generally, sufficient labeled data is a prerequisite for training accurate deep learning models. However, it is infeasible to label the continuously generated, massive IoT time series accurately. Second, IoT devices with limited storage and computing resources cannot support deep learning models with a large number of parameters \cite{mohammadi2018deep}. Complex models for hosts in traditional networks may fail on IoT devices. On the other hand, complicated calculations could generate a high computation overhead and lead to a rapid decrease in the lifetime of IoT devices. Other challenges stem from the understanding of DL models. For example, most deep neural networks operate as black boxes and offer little interpretability \cite{chakraborty2017interpretability}. Besides, different ML tasks require different capabilities of DL models, making it difficult to select the appropriate model that best adapts to a given problem. In this article, we systematically review unsupervised DL methods for IoT time series. Particularly, we investigate unsupervised anomaly detection and clustering under a unified framework to provide insights into the structures and capabilities of DL models.

\begin{figure*}[!t]
\centering
\includegraphics[width=7in]{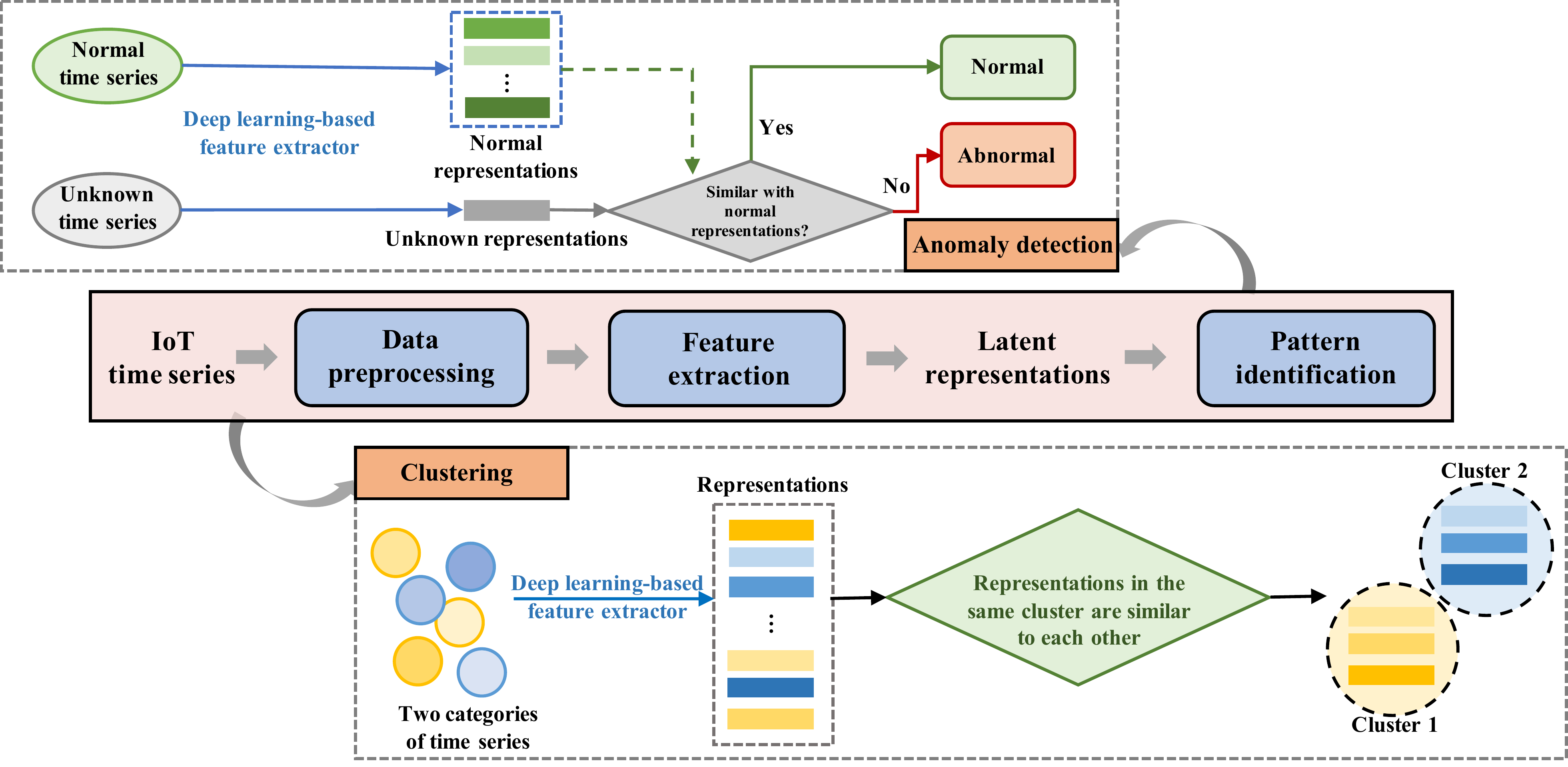}
\caption{A unified deep learning framework for unsupervised IoT time series analysis.}
\label{fig1.1}
\end{figure*}

\section{The unified framework of DL-based time series analysis}
\label{UniModel}
Time series analysis is of great importance for IoT system management. In this section, we investigate anomaly detection and clustering to summarize the general process of analyzing IoT time series using DL techniques, as shown in Fig. \ref{fig1.1}. Without loss of generality, this process can be abstractly divided into two core stages, namely feature extraction and pattern identification. The aim of feature extraction is essentially the same for different time series analysis tasks. That is, selecting an appropriate DL model according to the characteristic of input data so that the learned representations can describe the state of the monitored system as accurately as possible. But the second stage, pattern identification, varies from task to task and requires carefully designed objective functions. For example, anomaly detection aims to identify whether the features of an unknown sequence are similar to those of normal sequences. The clustering, on the other hand, aims to organize sequences with similar features into the same group. Specifically, based on the characteristics of IoT time series data, we introduce some special designs when modeling and processing IoT time series data in \ref{DataPreprocessing}. Then we discuss the other two important steps of IoT time series analysis, feature extraction, and pattern recognition in \ref{sub-FeatureExtraction} and \ref{PatternIdentification}, respectively.

\subsection{Data Preprocessing for IoT time series}
\label{DataPreprocessing}
As mentioned earlier, the challenges of IoT time series analysis mainly arise from the complex spatial temporal dynamics of the massive and high dimensional data. For multivariate time series, there are mainly five modeling strategies \cite{choi2021deep}: 1) using raw data as input directly; 2) extracting the main features via dimensional reduction; 3) using a 2D matrix to capture the relationships among individual variables directly; 4) constructing a graph to define an explicit topological structure and learn the causal relationship among individual variables; 5) defining correlations by specific distribution such as multivariate Gaussian distribution. 

There have also been designs for dealing with non-stationarity, heterogeneity, and noise of IoT time series. For non-stationary time series, researchers can convert them into stationary time series through methods such as detrending to reduce their damage to DL models.

The heterogeneity in IoT data sources limits learning techniques from realizing optimal performance. There have been studies regarding multimodal source fusion and heterogeneous data processing for IoT. Time series encoding is one of the techniques to mitigate the effect of IoT time series heterogeneity. \cite{abdel2020deep} encoded the time series data from different smartphone inertial sensors into three-channel image representation (i.e., RGB) to improve the accuracy of heterogeneous human activity recognition. Besides, some studies have shown that certain types of neural networks are robust to data heterogeneity. For example, in smart device localization, RSS values vary at the same location for different smart devices because of the difference in the receiver antenna, receiving circuit, frequency bands, and other factors. To cope with this problem caused by data heterogeneity, \cite{pandey2020residual} found that using the Residual Neural Network can reduce the localization error.

Using low-quality noisy IoT data as input can lead to incorrect analysis results. Therefore, denoising is a key preprocessing step for IoT time series analysis. Denoising methods for time series mainly include 1) mathematical transformations \cite{alrumaih2002time}, such as Fourier and wavelet transforms; 2) deep-learning-based supervised denoising \cite{frusque2022robust}, such as Denoising Autoencoder (DAE).

\begin{figure}[!t]
\centering
\subfigure[Deceleration]{
\label{fig3.2a}
\includegraphics[width=0.45\hsize]{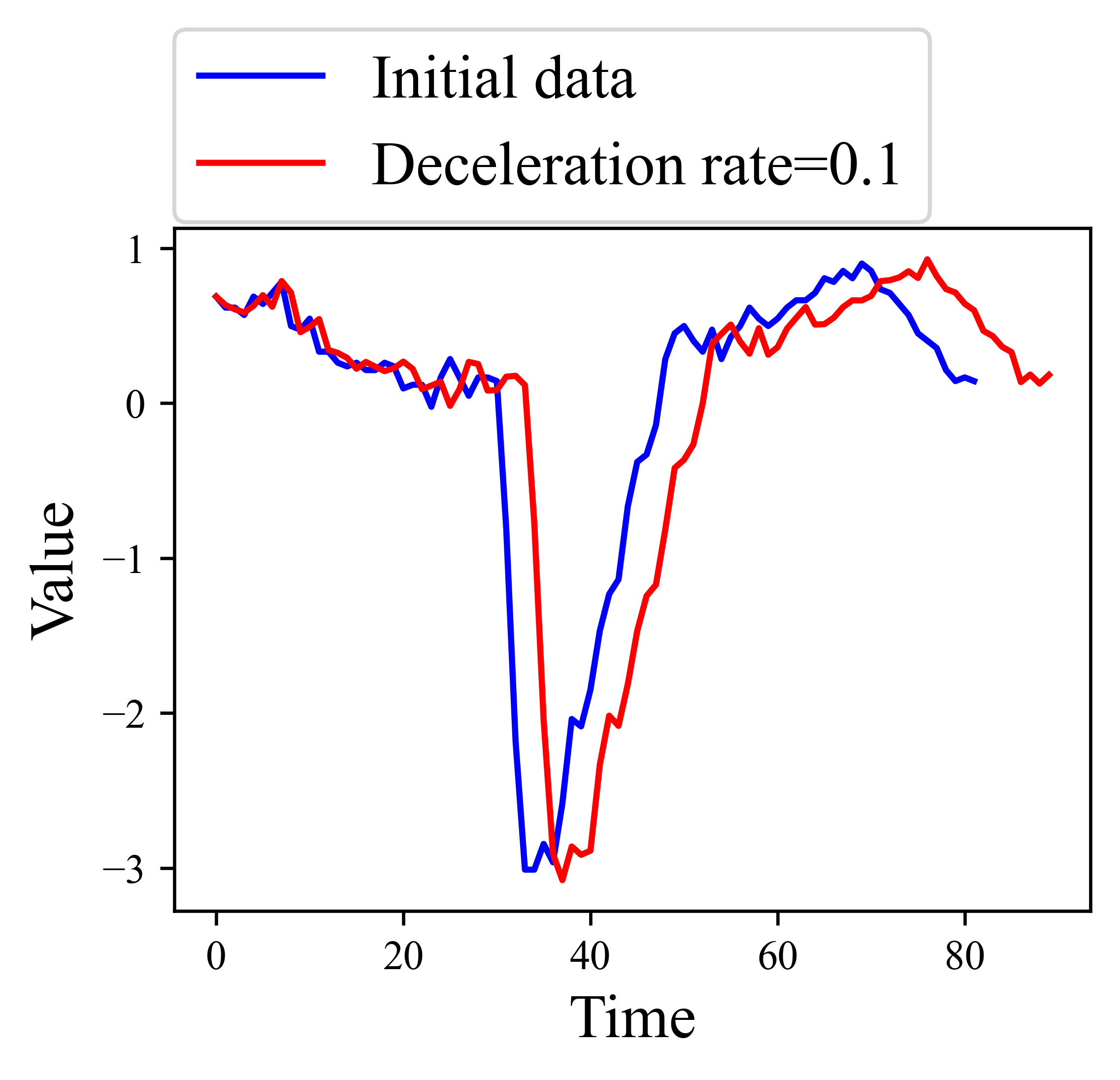}
}  
\subfigure[Acceleration]{
\label{fig3.2b}
\includegraphics[width=0.45\hsize]{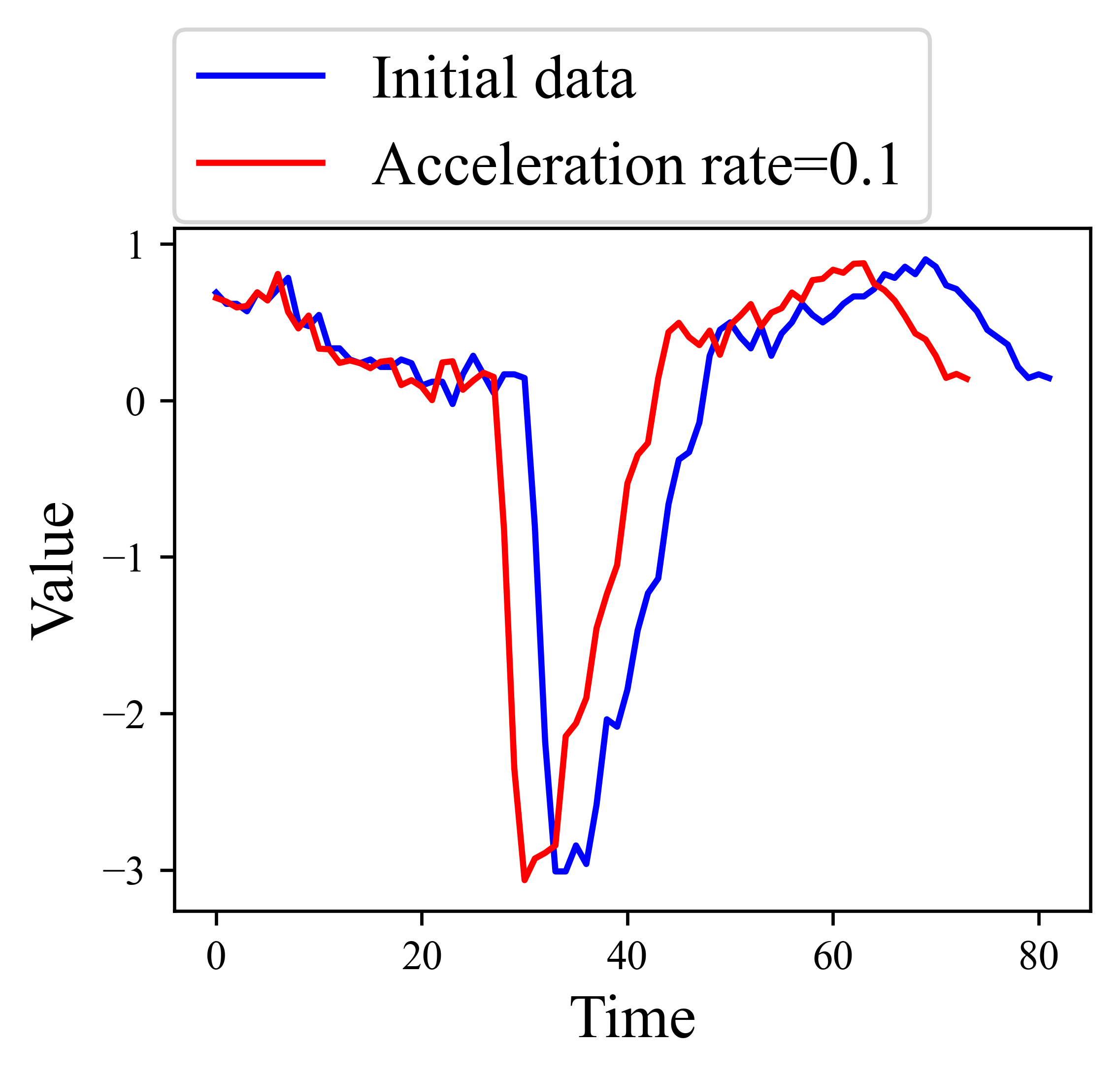}
} 
\hspace{2pt}

\subfigure[Right Rotation]{
\label{fig3.2c}
\includegraphics[width=0.45\hsize]{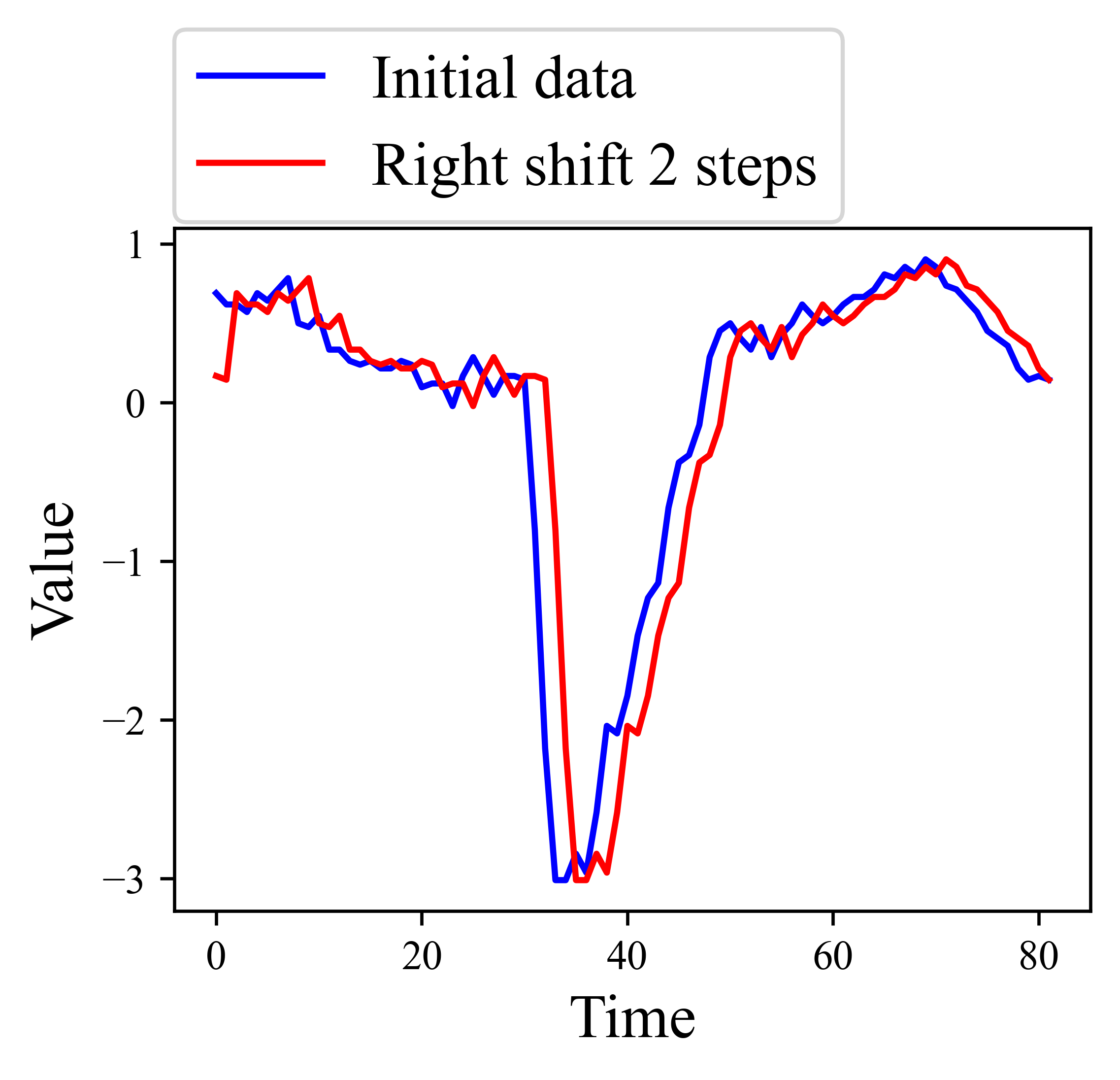}
}
\subfigure[Left Rotation]{
\label{fig3.2d}
\includegraphics[width=0.45\hsize]{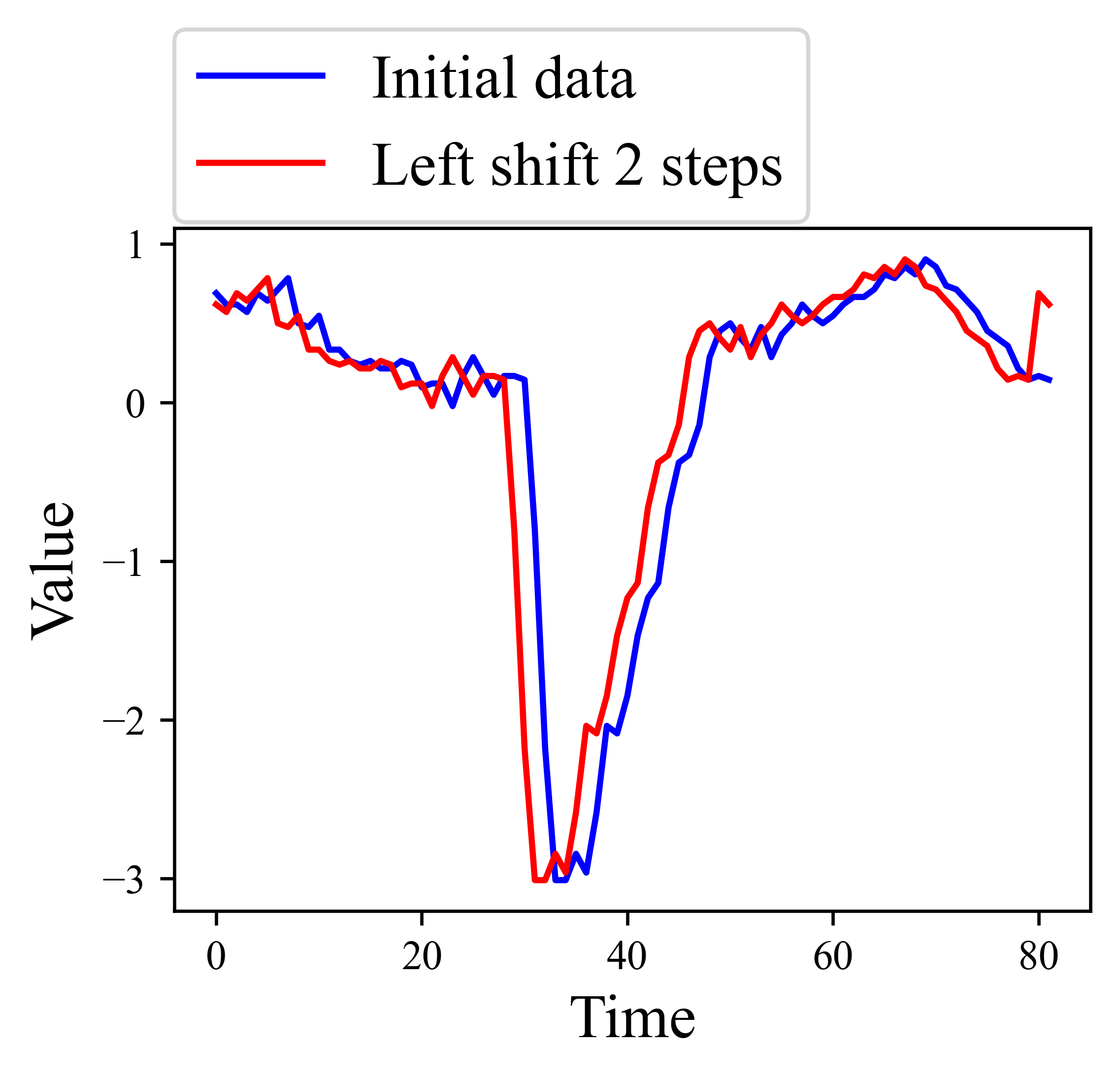}
}
\hspace{2pt}

\subfigure[Time Warping]{
\label{fig3.2e}
\includegraphics[width=0.45\hsize]{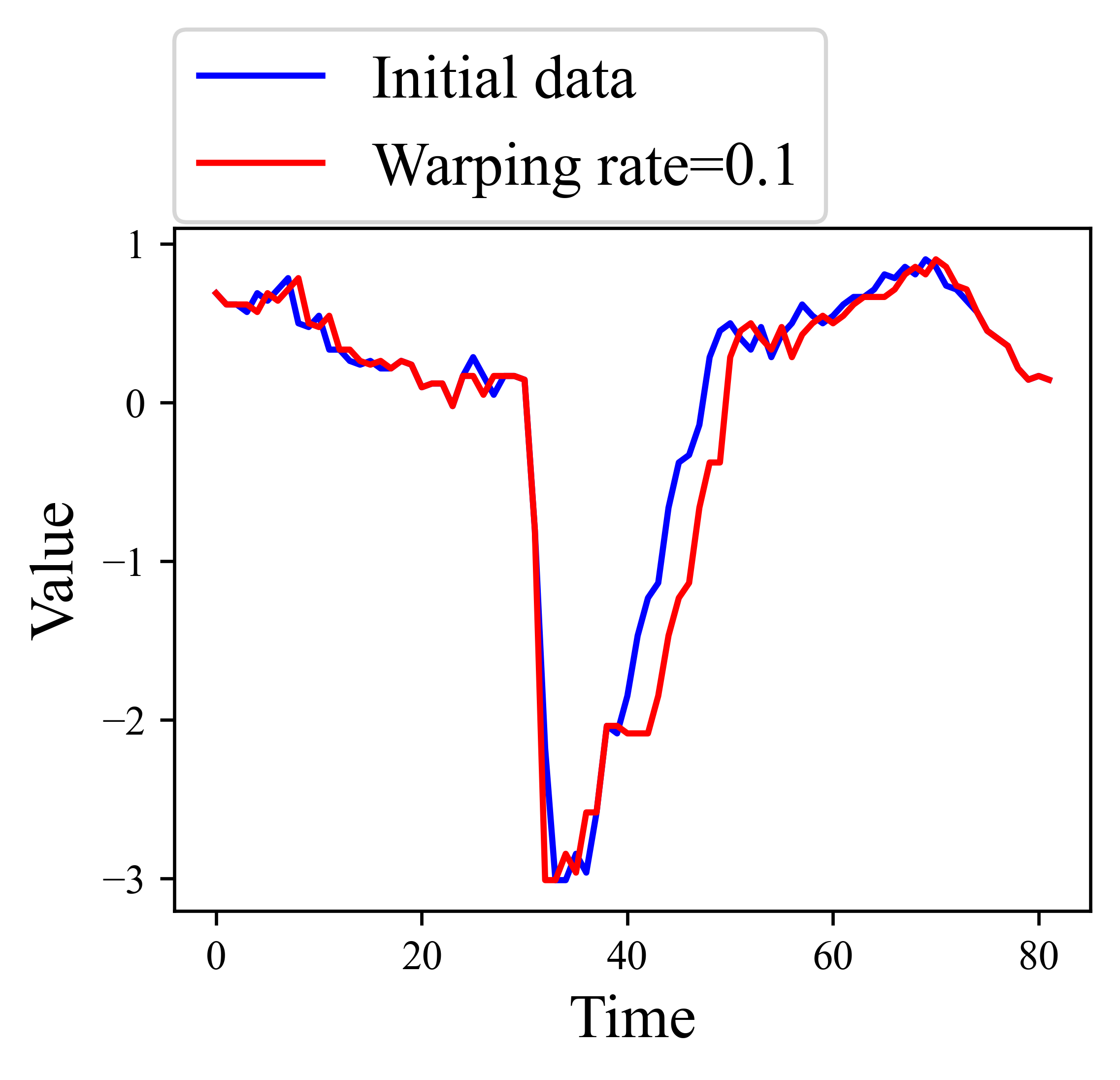}
}
\subfigure[Missing Value Simulation]{
\label{fig3.2f}
\includegraphics[width=0.45\hsize]{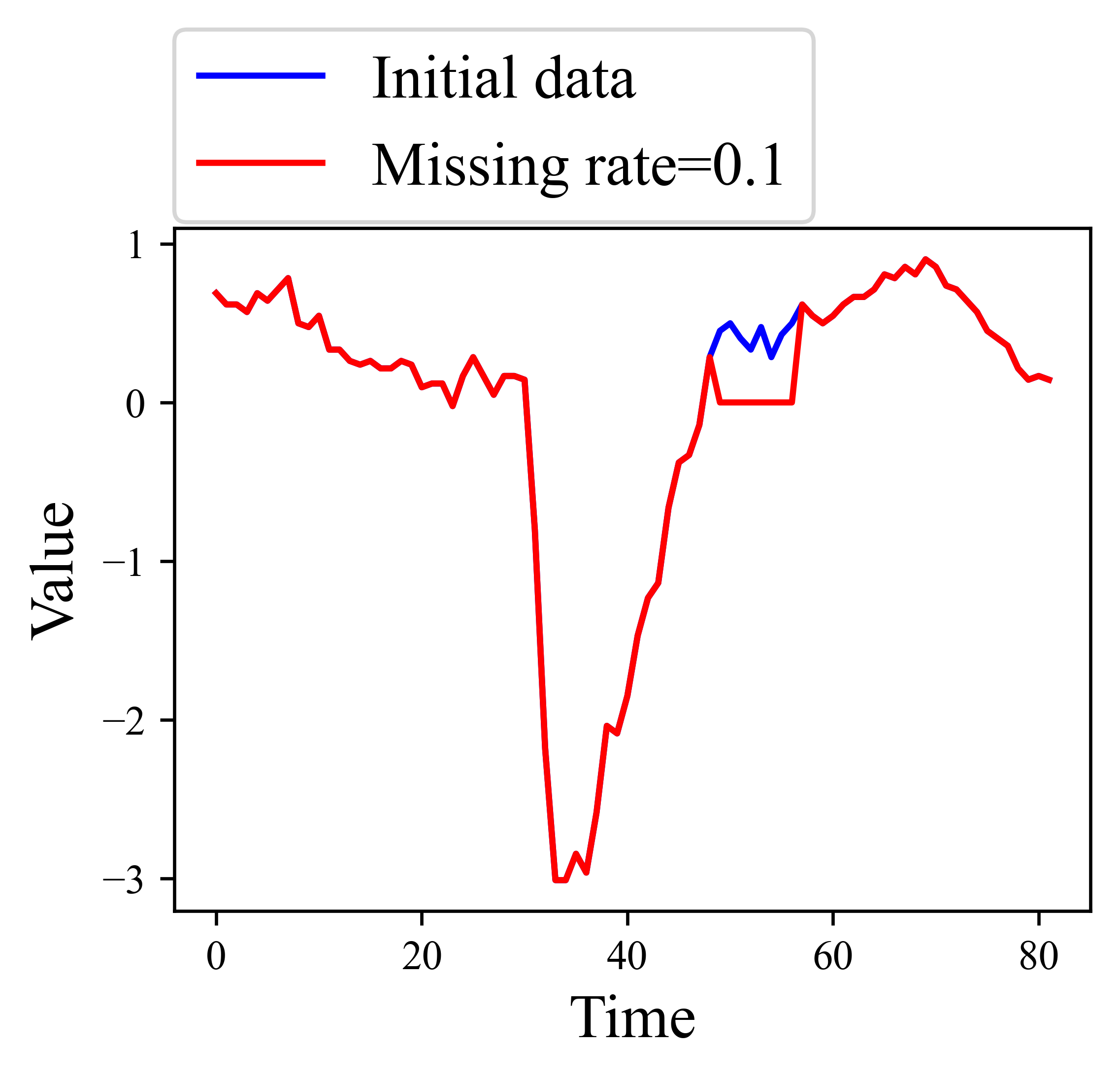}
}
\hspace{2pt}

\subfigure[Time Domain Noise]{
\label{fig3.2g}
\includegraphics[width=0.45\hsize]{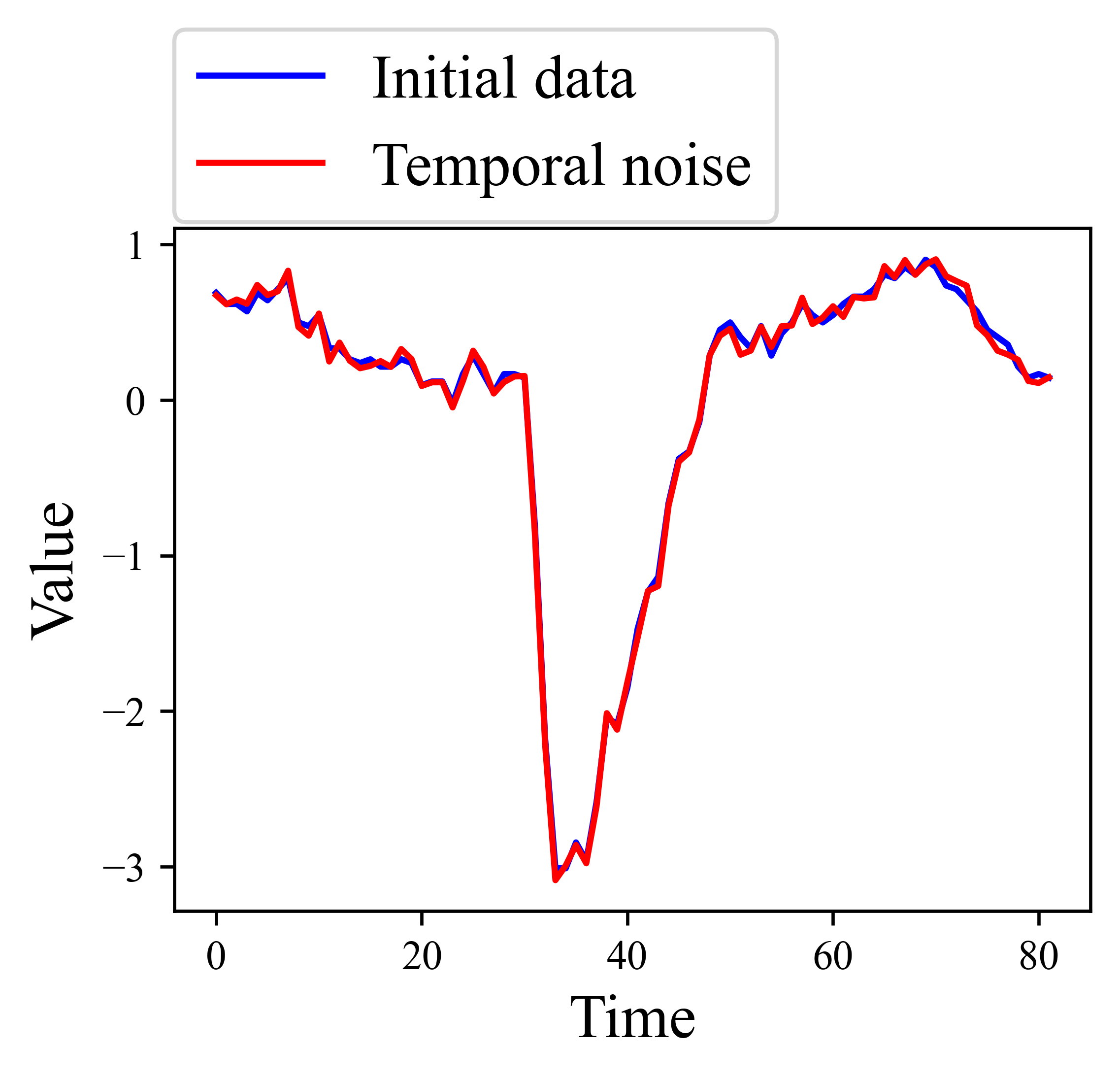}
}  
\subfigure[Frequency Domain Noise]{
\label{fig3.2h}
\includegraphics[width=0.45\hsize]{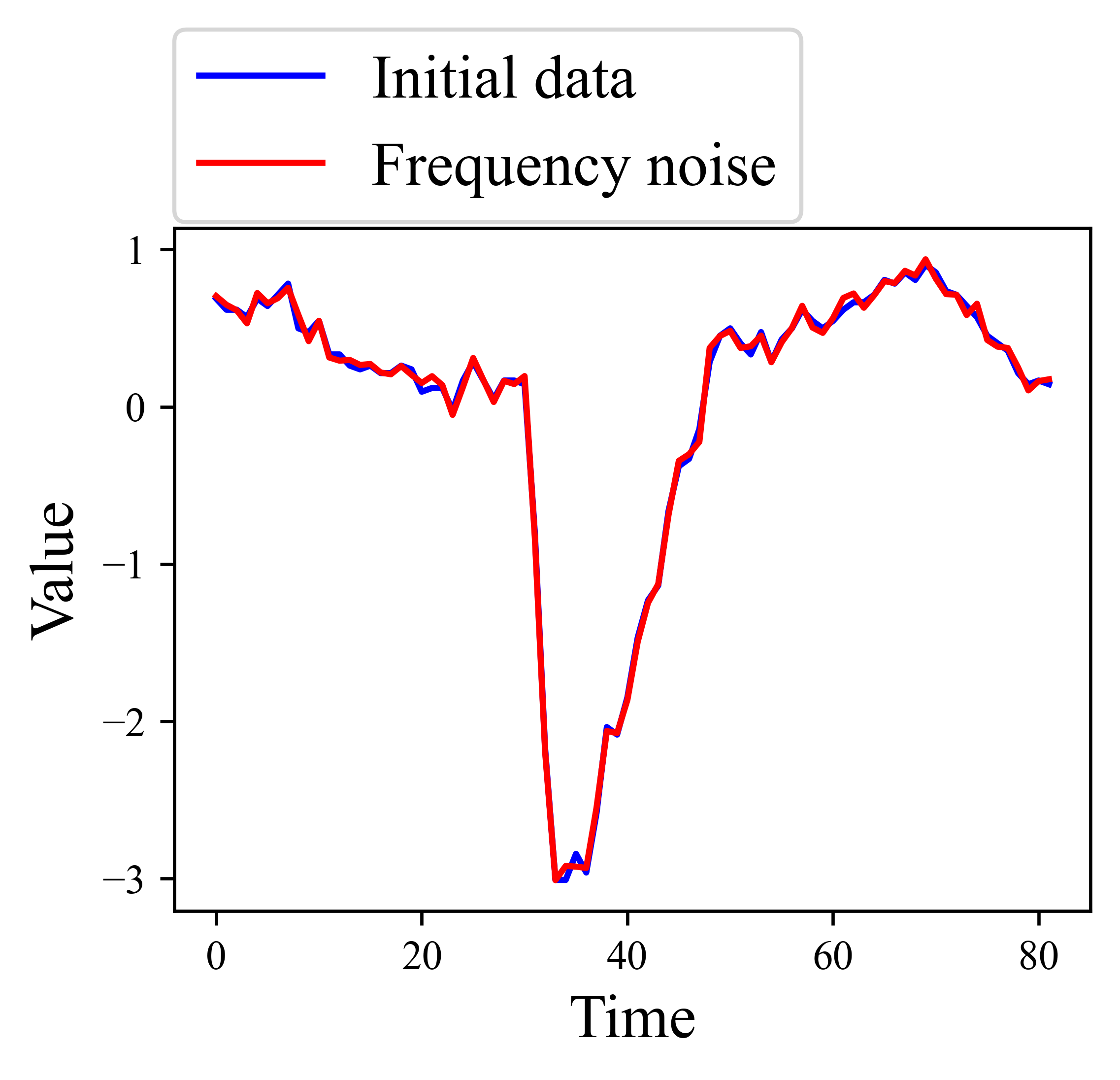}
} 
\caption{Visualization of several data augmentation methods for ECG series. The blue lines represent the original time series. The red lines represent the augmented time series. (a) Deceleration rate = 0.1. (b) Acceleration rate = 0.1. (c) Shift two steps to the right. (d) Shift two steps to the left. (e) Time warping rate = 0.1. (f) Missing rate = 0.1. (g) Add white Gauss noise ($\mu=0,\sigma=0.05$) in the time domain. (h) Add white Gauss noise ($\mu=0,\sigma=0.5$) in the frequency domain.}
\label{fig3.2}
\end{figure}

It is also worth noting that in some cases where the quantity or quality of IoT data is insufficient, data augmentation \cite{Goodfellow2015DeepL} can be applied to facilitate the training of DL models. Generally, data augmentation increases the amount of data by adding synthetic data or slightly modified copies of existing data. Looking at time series data, basic approaches to data augmentation include time domain methods, frequency domain methods, and time-frequency domain methods \cite{Wen2020TimeSD}. Six commonly used augmentation methods for ECG series \cite{Dau2019TheUT}, a kind of IoT time series, are introduced in Fig. \ref{fig3.2}: speed variation, rotation, time warping, missing value simulation, adding noise in time-domain, and adding noise in frequency domain.

\begin{itemize} 
\item \emph{Speed Variation.} Resample the time series to simulate different heart rates. The deceleration and the acceleration of ECG time series are shown in Fig. \ref{fig3.2a} and Fig. \ref{fig3.2b}, respectively.

\item \emph{Rotation.} Time series data often contains synchronization errors, or the timestamps are not perfectly aligned. Rotating the time series can simulate synchronization errors. The right rotation and the left rotation of ECG time series are shown in Fig. \ref{fig3.2c} and Fig. \ref{fig3.2d}.

\item \emph{Time Warping.} This method can be used to simulate the immediate increase or decrease of heart rate. Remove $r\%$ points at random from the time series, and then add $r\%$ points at random. A case of $r$=10 is  shown in Fig. \ref{fig3.2e}.

\item \emph{Missing Value Simulation.} Randomly set the value of a subsequence to 0 to simulate the missing values caused by equipment failure. A case with 10\% missing data points is shown in Fig. \ref{fig3.2f}.

\item \emph{Time Domain Noise.} Add noise to the time series in the time domain. A time series with Gaussian noise is shown in Fig. \ref{fig3.2g}.

\item \emph{Frequency Domain Noise.}
In low signal-to-noise ratio (SNR) conditions, adding noise directly in the time domain may destroy data characteristics. In contrast, adding noise in the frequency domain is a better choice. A time series that contains Gaussian noise in the frequency domain is shown in Fig. \ref{fig3.2h}.

\end{itemize}

\subsection{Feature Extraction}
\label{sub-FeatureExtraction}
Conversion of given input data into set of features are known as Feature Extraction \cite{dara2018feature}. Features learned from the initial dataset are expected to be descriptive and non-redundant, simplifying subsequent analysis. Unlike traditional handcrafted feature extraction, DL can automatically learn time series features through complex nonlinear transformations.

Formally, a feature extractor $f(\cdot)$ can encode a univariate time series ${X}=\{{x}_{t}\}_{{t}\in{T}}$ into a latent vector $Z$ or encode a multivariate time series $\boldsymbol{X}=\{\boldsymbol{x}_{t}\}_{{t}\in{T}}$ into a latent matrix $\boldsymbol{Z}$. Using $Z$ or $\boldsymbol{Z}$ for downstream tasks can reduce the cost of feature analysis and improve the accuracy and efficiency of models. We discuss how to select an effective feature extraction model according to the characteristics of the input time series in subsection \ref{FeatureExtraction}.

\subsection{Pattern Identification}
\label{PatternIdentification}
Well-known time series analysis tasks include anomaly detection, clustering, classification, and forecasting. Each task identifies a particular data pattern from the learned representations. For example, a semi-supervised anomaly detection model learns the pattern of normal data. Then test samples that deviate from the normal pattern will be deemed as anomalies. As for clustering, the DL model explores different patterns in a dataset so that samples conforming to the same pattern will be grouped into the same cluster. 

In the pattern identification phase, some works apply traditional algorithms such as anomaly detection algorithms (Isolation Forest \cite{liu2008isolation}, one-class SVM \cite{li2003improving}, etc.) or clustering algorithms (K-Medoids, K-Means, etc.) to the previously learned representations. However, in this case, the feature extraction and the pattern identification are two independent stages, leading to suboptimal performance. A better strategy is to analyze problems end-to-end using DL models, that is, to automatically learn data features and calculate the results. Next, we will categorize and comment on DL-based anomaly detection and clustering methods.

\begin{figure*}[!t]
\centering
\includegraphics[width=5.5in]{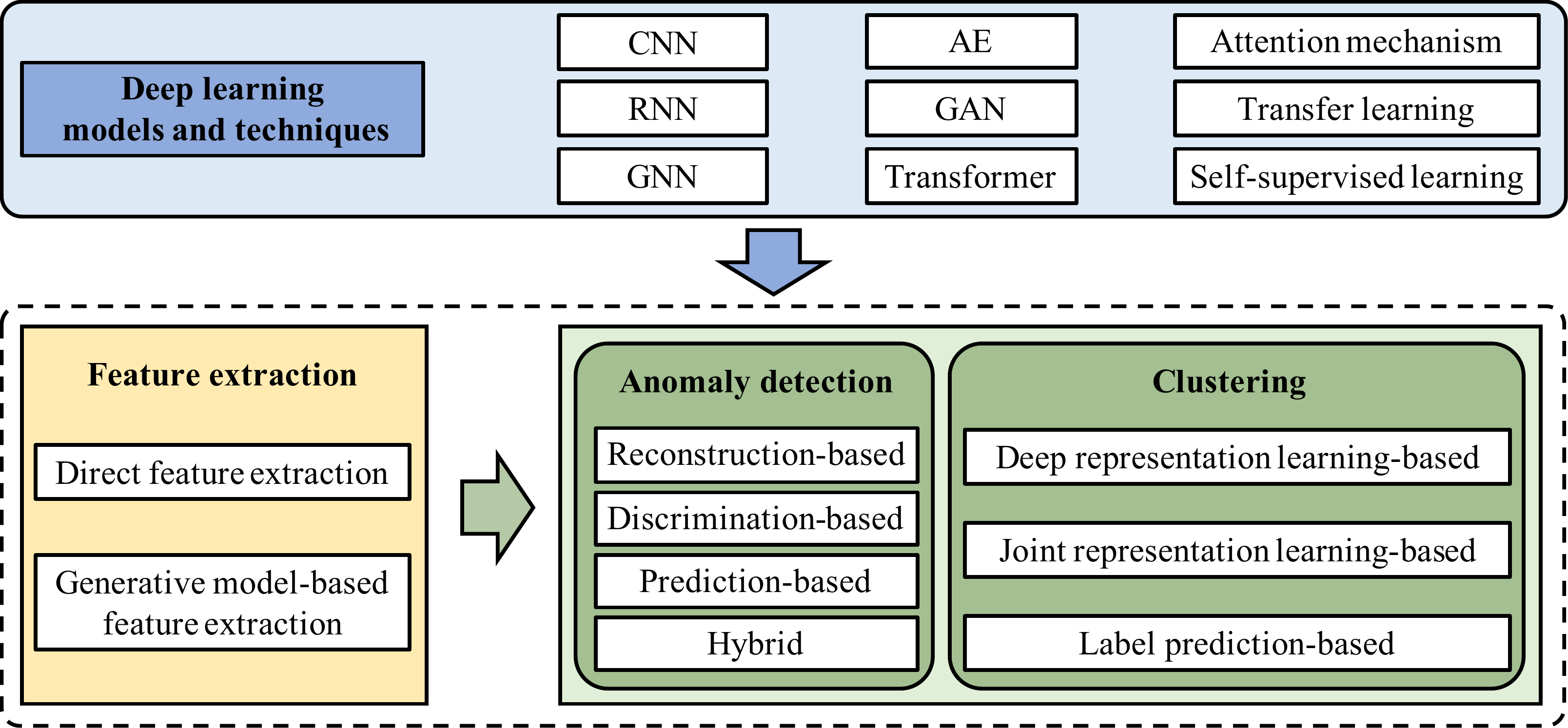}
\caption{Taxonomies of unsupervised DL methods for time series anomaly detection and clustering.}
\label{fig4_main}
\end{figure*}

\section{Models based on Deep Learning}
\label{deepmodels}
In this section, we first introduce several commonly used DL models and techniques. Then we review and categorize current DL-based methods for time series feature extraction, anomaly detection, and clustering.

\subsection{Deep Learning Models and Techniques}
The most basic model of artificial neural network (ANN) is multi-layer perceptron (MLP), which is a fully connected feed-forward neural network \cite{Haselsteiner2000UsingTN}. Many types of neural networks have been proposed in the past decade. This subsection introduces DL models commonly used for time series analysis.

\subsubsection{Convolutional Neural Network (CNN)} CNN is suitable for processing data with a grid-like structure \cite{Goodfellow2015DeepL}. It is a feed-forward neural network composed of three types of layers: convolutional layers, pooling layers, and fully connected layers. The convolutional layer is the core building block of CNN. It involves the multiplication of the input with a set of weights called a filter or a kernel. In the convolutional layers, the previous layer's output will be used as the input of the next layer.

\subsubsection{Recurrent Neural Network (RNN)} 
RNNs are dominant in research areas involving sequential data. The typical architecture of RNN is a cyclic connection that enables the RNN to update the current state based on past states and current input data \cite{yu2019review}. However, gradients of conventional RNNs may tend to disappear or explode during the propagation, which makes them difficult to learn long-term dependencies. Long Short-Term Memory (LSTM) \cite{Hochreiter1997LongSM} and Gated Recurrent Units (GRUs) \cite{Chung2014EmpiricalEO} are proposed to resolve this issue. They use internal mechanisms called gates to regulate the flow of information.

There have been some attempts to combine RNNs and CNNs. Convolutional LSTM (ConvLSTM) \cite{shi2015convolutional} replaces the fully-connected layers in LSTM with convolutional layers to capture the spatio-temporal correlation. The Quasi-RNN model \cite{bradbury2016quasi} alternates convolutional layers and simple recurrent layers to allow parallel processing. The Dilated RNN \cite{chang2017dilated} uses dilated recurrent skip connections to reduce model parameters and improve training efficiency. Bai et al. \cite{Bai2018AnEE} proposed a general architecture of convolution-recurrent models named Time Convolutional Network (TCN). They used causal convolution to fit sequential data and extended convolution and residual modules to memorize past states.

\subsubsection{Graph Neural Networks (GNN)}
GNNs \cite{gori2005new} emerge as new approaches for modeling graph-structured data. There are usually complex topological relationships between sensors in IoT scenarios, so the whole system can be seen as a graph structure where nodes represent sensors and edges describe the relationships among nodes \cite{chen2021learning, wu2020connecting}. GNNs have proved to be effective for large-scale multi-relational data modeling \cite{schlichtkrull2018modeling}, making them promising for modeling high-dimensional time series. Up till now, GNNs can be categorized into Recurrent GNNs, Convolutional GNNs, Graph Autoencoders, and Spatial Temporal GNNs \cite{wu2020comprehensive}.  

\subsubsection{Autoencoder (AE)} 
Autoencoders are primarily designed to encode an input into a latent representation and then reconstruct it \cite{bank2020autoencoders}. Generally, the dimension of the encoded representation is smaller than the input dimension. The simplest form of an autoencoder is a feedforward, non-recurrent neural network that employs an input layer and an output layer connected by one or more hidden layers. 

\subsubsection{Generative Adversarial Network (GAN)}
GANs are machine learning frameworks consisting of two neural networks that compete with each other: one (the generator $G$) is trained to generate fake data, and the other (the discriminator $D$) is trained to discern the fake data from the real one. $G$ generates better data during training, while $D$ becomes more skilled at discerning fake data. GANs can implicitly model the high-dimensional distribution of data \cite{creswell2018generative}.

\subsubsection{Transformer} 
Transformer \cite{vaswani2017attention} is an architecture that leverages the attention mechanism to process sequence data. Unlike RNNs, which rely on an inherently sequential nature, Transformer allows the model to access any part of the history regardless of distance, making it significantly more parallelizable and potentially more suitable for capturing long-term dependencies. Canonical Transformer follows an encoder-decoder structure using stacked self-attention and point-wise, fully connected layers. Li et al. \cite{li2019enhancing} improved the canonical Transformer to make it more ideal for time series modeling. Precisely, they used convolutional self-attention to utilize local context and proposed LogSparse self-attention to break the memory bottleneck.

Next we introduce techniques that can facilitate DL-based time series analysis, such as attention mechanism, transfer learning, and self-supervised learning.

\setcounter{subsubsection}{0} 
\subsubsection{Attention Mechanism} 
The attention mechanism \cite{vaswani2017attention} is a component of the neural network architecture responsible for capturing the correlations between different parts of data. It helps the model automatically identify the crucial parts of the input data and assign them large weights. Attention mechanism has been widely used in many fields, such as machine translation, speech recognition, and image caption. In the field of time series modeling, it is assumed that previous time steps have different correlations with the current state. Thus, models with attention mechanisms can adaptively select appropriate previous time steps and aggregate the information to form a refined output \cite{Zhang2019ADN,Ienco2020DeepMT,Carrasco2019AnUF}. 

\subsubsection{Transfer Learning (TL)} 
It is challenging to train a reliable DL model through the traditional supervised learning paradigm with insufficient labeled data. To cope with this problem, TL utilizes labeled data from different but related tasks to facilitate the learning of the target task. In other words, the knowledge learned from a related task is transferred to the target task \cite{torrey2010transfer}. Generally, the target model is first pre-trained on an auxiliary dataset and then fine-tuned on target data. Wen et al. \cite{Wen2019TimeSA} used TL to improve their time series anomaly detection model's generalization capabilities for unknown anomalies. They synthesized a pre-training dataset with short-term, medium-term, and long-term anomalies. These anomalies can be considered components of other complex anomalies, so the information learned from the synthetic dataset would benefit general anomaly detection tasks.

\subsubsection{Self-Supervised Learning (SSL)} 
Another commonly used strategy to deal with the lack of labeled data is SSL. As a branch of unsupervised learning, SSL leverages input data itself as supervision \cite{Liu2020SelfsupervisedLG}. The general process of SSL consists of two steps. The first step is to train a pretext task on a large amount of unlabeled data. Then the second step is to fine-tune the pre-trained model according to the target task. Ma et al. \cite{Ma2019LearningRF} used SSL to improve representation learning. First, they generated a fake sample for each unlabeled time series by shuffling partial timestamps. Then, the classification of true and fake samples was used as an auxiliary task.

\subsection{Models for Feature Extraction}
\label{FeatureExtraction}

\begin{table*}[t]
\renewcommand{\arraystretch}{1.2} 
\caption{Summary of time series feature extraction methods based on DL models.}
\label{table4.1}
\centering
\begin{tabular}{p{0.17\textwidth}<{\centering}|c|l}
\hline \hline
Category & Major Model & Related Works\\
\hline
\multirow{5}{0.17\textwidth}{\centering Direct Feature Extraction}
& \multirow{2}{*}{RNN} 
& \cite{Ienco2020DeepMT} (Bi-direction GRU), \cite{Walton2017UnsupervisedAD} (LSTM), \cite{ Inoue2017AnomalyDF} (LSTM), \\ 
&& \cite{Trosten2019RecurrentDD} (Bi-direction GRU), \cite{Hundman2018DetectingSA} (LSTM), \cite{Su2019RobustAD} (GRU) \\
\cline{2-3}
& CNN & \cite{Zhang2019ADN, Carrasco2019AnUF, Ren2019TimeSeriesAD}\\
\cline{2-3}
& RNN + CNN  & \cite{liu2019deep}\\
\cline{2-3}
& GNN & \cite{deng2021graph} (GAT), \cite{chen2021learning} (Graph convolution), \cite{zhao2020multivariate} (GAT)\\
\hline
\multirow{6}{0.17\textwidth}{\centering Generative Model-Based Feature Extraction} 
& \multirow{6}{*}{Autoencoder} 
& \cite{Ma2019LearningRF} (Bi-direction dilated recurrent autoencoder), \cite{Zhang2019ADN} (Convolutional autoencoder),\\
&& \cite{Ienco2020DeepMT} (Bi-direction recurrent autoencoder), \cite{Carrasco2019AnUF} (Convolutional autoencoder), \\
&& \cite{Pereira2018UnsupervisedAD} (Bi-direction DAE), \cite{Lee2017AnomalyDI} (SAE), \cite{Tavakoli2020ClusteringTS} (SAE), \cite{Lin2020AnomalyDF} (VAE), \\
&& \cite{Biradar2019ChallengesIT} (Convolutional autoencoder),
\cite{Richard2020AutoencoderbasedTS} (Convolutional autoencoder),
\cite{Kalinicheva2020UnsupervisedSI} (Convolutional autoencoder),\\ 
&& \cite{Bhatnagar2017UnsupervisedLO} (Convolutional autoencoder and recurrent autoencoder),
\cite{Meng2020ATC} (Time convolutional autoencoder) \\ 
\cline{2-3}
& GAN & \cite{Li2019MADGANMA} (Recurrent GAN), \cite{Bashar2020TAnoGANTS} (GAN)\\
\hline
\end{tabular}
\end{table*}

A direct strategy of time series feature extraction is to input time series into a feed-forward neural network and take the output as representations. The most basic neural networks are MLPs. But MLPs ignore the temporal dependencies of time series. Researchers have applied various more complex neural networks to feature extraction of time series.

RNNs can capture complex temporal dependencies between different time steps through a cyclic connection. An RNN trained on normal data can learn the normal behavior of a dynamic system \cite{Walton2017UnsupervisedAD}, \cite{Inoue2017AnomalyDF}. However, it is difficult for conventional RNNs to learn long-term dependencies due to the vanishing gradient. By utilizing the mechanism of gates, LSTMs perform better at modeling long sequences, but the multiplying parameters of LSTMs increase the risk of overfitting. GRUs are more suitable for small-scale data because they can achieve similar performance as LSTMs with simpler structures and fewer parameters. Also noteworthy, the state of an RNN is usually passed from the front to back. But in some cases, modeling time series simultaneously from forward and reverse can utilize more context information and promote representation learning. For example, Trosten et al. \cite{Trosten2019RecurrentDD} used a bidirectional GRU to learn representations of multivariable time series.

CNNs can extract patterns of high-dimensional data with complex structures \cite{Gorokhov2017ConvolutionalNN}. Zhang et al. \cite{Zhang2019ADN} and Carrasco et al. \cite{Carrasco2019AnUF} used CNN to extract representations of the sequences of system feature maps. Ren et al. \cite{Ren2019TimeSeriesAD} transformed time series into saliency maps by the Spectral Residual (SR) algorithm and then analyzed the maps with CNNs. 

Liu et al.\cite{liu2019deep} used a hybrid model of LSTM and CNN to capture the long-term and short-term dependencies in the network traffic sequence. First, they built an LSTM model on the network traffic to capture long-term dependencies. Then a CNN is applied to the hidden states of the LSTM to extract the local spatial information. The final representations learned by the hybrid model fully encapsulate the sequence characteristics.

When analyzing multivariate time series, the most straightforward strategy is to treat each dimension as an independent univariate time series. For example, Hundman et al. \cite{Hundman2018DetectingSA} constructed an LSTM-based predictor for each univariate time series when modeling multivariate aerospace remote sensing data. However, treating each dimension separately has the following disadvantages. First, it is labor-intensive to train and maintain an separate model for each dimension. Besides, the dependencies among multiple dimensions are ignored. Methods such as multivariate Gaussian distributions \cite{ding2019real} and two-dimensional matrices \cite{choi2021deep} can model these dependencies explicitly.

Another promising approach to deal with multivariate time series is GNN. Deng et al. \cite{deng2021graph} encoded the asymmetric relationships between pairs of sensors as directed edges in a graph. In the case without prior information, the graph's adjacency matrix is obtained by similarity measurement and embedded by a Graph Attention Network (GAT). Chen et al. \cite{chen2021learning} devised a directed graph structure learning policy to automatically discover the adjacency matrix. Then graph convolution layers integrated with different level dilated convolution layers are utilized to capture hierarchical temporal context. Zhao et al. \cite{zhao2020multivariate} modeled the inter-feature correlations and temporal dependencies of multivariate time series with two GATs in parallel, followed by a GRU network to capture long-term dependencies.  

\emph{Generative Model-Based Methods:}
In addition to the above works, some researchers consider using generative models to achieve better representation capabilities. Generative models assume that the available data is generated by some unknown distribution and try to estimate this distribution. Commonly used generative models include autoencoders and GANs. 

\begin{figure}[!t]
\centering
\includegraphics[width=3.2in]{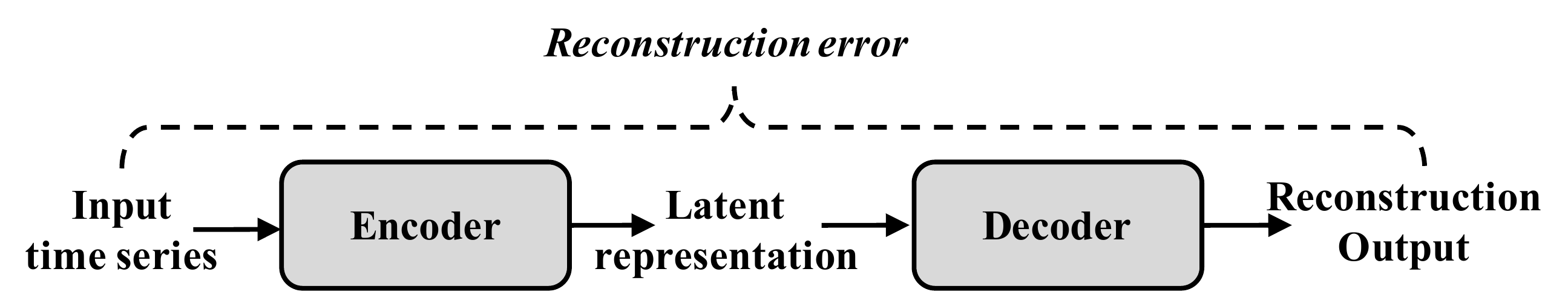}
\caption{Feature extraction based on autoencoders.}
\label{fig4.1}
\end{figure}

Autoencoders learn the distribution of the input data by minimizing the reconstruction loss that measures the distance between the output of the decoder and the input data. As shown in Fig. \ref{fig4.1}, the output of the encoder can be regarded as a meaningful representation that retains important patterns of the input data. The simplest autoencoder consists of three layers: an input layer, a hidden layer, and an output layer. Further, a stacked autoencoder (SAE)\cite{chen2014deep} contains multiple hidden layers, where the output of the front hidden layer is used as the input of the next hidden layer. Each layer produces a more abstract representation than the one before because the representation is obtained by composing more operations \cite{vincent2008extracting}. Doyup Lee \cite{Lee2017AnomalyDI} used an SAE with three hidden layers to detect anomalies in a database management system. Tavakoli et al. \cite{Tavakoli2020ClusteringTS} used an SAE to cluster financial data. The number of hidden layers and hidden units is determined by the dataset.

However, traditional autoencoders are trained only to minimize the reconstruction errors, which may lead the model to copy the input without learning helpful information. Denoising autoencoders (DAEs) are proposed to solve this problem. A DAE takes noisy data as input but is forced to reconstruct the clean version of the input \cite{Carrasco2019AnUF, Pereira2018UnsupervisedAD}. Another kind of well-known autoencoders, Variational autoencoders (VAEs) \cite{Kingma2014AutoEncodingVB}, attempt to model the underlying probability distribution of data. The latent space of a VAE is forced to obtain a specified distribution so that a random vector sampled from the latent distribution can generate meaningful content similar to the real data. The latent constraint also improves the generalization ability of VAEs. Lin et al. \cite{Lin2020AnomalyDF} used VAEs to learn robust local features of time series windows.

Autoencoders can be constructed based on different neural networks. RNNs can enhance the sequence modeling capability of autoencoders. Ienco and Interdonato \cite{Ienco2020DeepMT} proposed a bidirectional GRU-based autoencoder to capture the complex temporal dependencies among multivariable time series. Ma et al. \cite{Ma2019LearningRF} proposed an autoencoder based on bidirectional extended RNNs \cite{Chang2017DilatedRN}. Compared with traditional RNNs, extended RNNs contain the multi-resolution extended skip connection that can reduce parameters, improve training efficiency, and maintain multi-level dependencies.

CNNs can enhance the ability of autoencoders to extract complex features. The encoder of a convolutional autoencoder (CAE) contains convolutional layers, and the decoder contains deconvolution layers. Biradar et al. \cite{Biradar2019ChallengesIT} used CAEs to extract features of traffic videos in their anomaly detection framework. Richard et al. \cite{Richard2020AutoencoderbasedTS} used CAEs to learn meaningful representations of electric power consumption time series. However, these CNN-based works ignore the temporal dependencies of time series data.

\cite{Zhang2019ADN} and \cite{Carrasco2019AnUF} proposed spatio-temporal autoencoders to deal with multivariable time series in Cyber-Physical-Systems (CPSs). They used 2D-CAEs to encode the correlations between different sensors and then used ConvLSTMs to capture the dynamic patterns of the system. Kalinicheva et al. \cite{Kalinicheva2020UnsupervisedSI} extracted features of the satellite image time series (SITS) with a 3D-CAE, in which the 3D filter could preserve the temporal dependencies between data. Compared with the ConvLSTM structure that combines 2D convolution and RNNs, the 3D convolution has a lower computational overhead when modeling sequence data.

Meng et al. \cite{Meng2020ATC} captured features of Cyber-Physical-Social Systems (CPSS) time series with a time convolutional network-based automatic encoder (TCN–AE). It's composed of causal convolution, dilated convolution, a residual module, and an FCN. TCN's parallelism and low memory overhead enable TCN-AE to be applied in large and complex systems in cloud-fog-edge computing. Bhatnagar et al. \cite{Bhatnagar2017UnsupervisedLO} considered different temporal resolutions when learning representations of human activities of varying time spans in videos. They used a set of CAEs to learn frame-level representations with varying intervals. Then multiple LSTM autoencoders are constructed on the output of CAEs to obtain the final representations.

\begin{figure}[!t]
\centering
\includegraphics[width=3.2in]{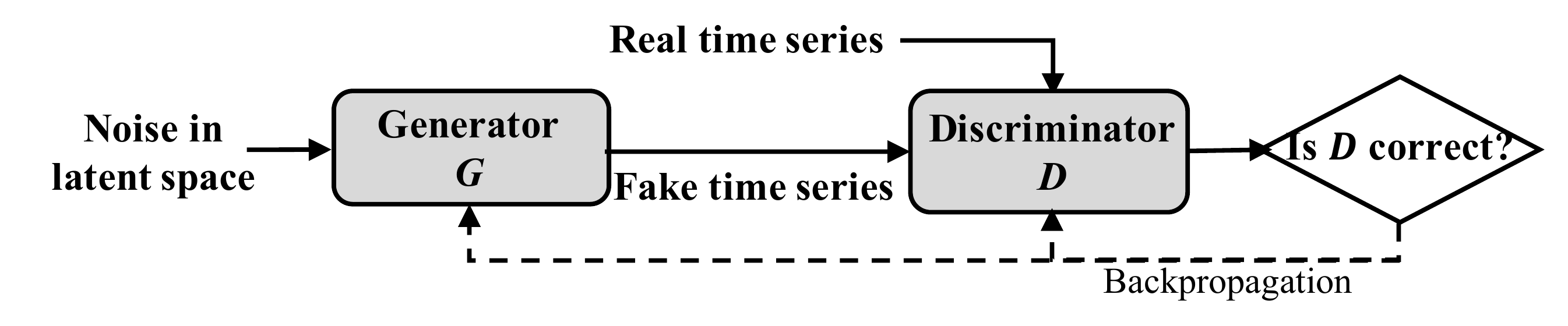}
\caption{Feature extraction based on GANs.}
\label{fig4.2}
\end{figure}

Besides autoencoder, another commonly used generative model is Generative Adversarial Network (GAN). \cite{esteban2017real} has proven that a trained GAN can model the distribution of high-dimensional time series data well. Bashar et al. \cite{Bashar2020TAnoGANTS} learned the distribution of given time series through adversarial training as Fig. \ref{fig4.2}. They simultaneously trained an LSTM-based generator $G$ that generates fake time series data and an LSTM-based discriminator $D$ to distinguish between the generated and real data. Li et al. \cite{Li2019MADGANMA} used LSTM-based GAN to learn the representations of multivariate time series in CPS systems. The proposed GAN framework processes time series from multiple sensors simultaneously to capture the latent interactions among the sensors. By being able to generate realistic data, the generator $G$ will have captured the hidden distributions of the training sequences. 

\subsection{Models for Anomaly Detection}
\label{ADTask}
According to the detection principles, existing DL-based time series anomaly detection methods can be divided into three categories: reconstruction-based methods, discrimination-based methods, and prediction-based methods.

\subsubsection{Reconstruction-Based Methods}
Reconstruction models refer to those neural networks with a bidirectional mapping between data space and latent space, such as autoencoders. The encoder maps the input data into a latent space, and the decoder maps the latent vector back to the data space. A reconstruction model trained on normal samples could learn the manifold of normal data. Therefore, a normal sample in the test set will be reconstructed well, but anomalous samples will not since the model has not seen anomalies during the training phase. In other words, it is reasonable to suspect that a sample of low reconstruction quality is abnormal.

\begin{table*}[t]
\renewcommand{\arraystretch}{1.2} 
\caption{Summary of DL-based time series anomaly detection methods.}
\label{table4.2}
\centering
\begin{tabular}{c|c|m{0.73\textwidth}}
\hline \hline
Work & Category & Description\\
\hline 
\cite{Zhang2019ADN}	& Reconstruction-based & Reconstruction errors of multi-scale and multivariable time series are used to detect and diagnose anomalies in complex systems. \\
\hline
\cite{Pereira2018UnsupervisedAD} & Reconstruction-based & A VAE is used to reconstruct the distribution parameters of the input data. The reconstruction probability is used as the anomaly score. \\
\hline
\cite{Lee2017AnomalyDI} & Reconstruction-based & Reconstruction errors of an autoencoder are used as the anomaly score.\\
\hline 
\cite{Bashar2020TAnoGANTS} & Reconstruction-based & Reconstruction errors of a GAN's generator are used as the anomaly score.\\
\hline 
\cite{Malhotra2016LSTMbasedEF} & Reconstruction-based & Reconstruction errors of an LTSM-based autoencoder are used as the anomaly score. \\
\hline
\cite{Audibert2020USADUA} & Reconstruction-based & A weighted sum of the reconstruction errors of two adversarially-trained autoencoders is used as the anomaly score.\\
\hline 
\cite{zhou2021feature} & Reconstruction-based & It learns representations more meaningful for anomaly detection through the process of reconstruction.\\
\hline 
\cite{Li2019MADGANMA} & Discrimination-based & The discrimination results and reconstruction residuals of GAN are combined as a novel anomaly score.\\
\hline
\cite{chen2021learning} & Prediction-based & Combine the graph structure with Transformer to model multivariable time series. \\
\hline
\cite{Walton2017UnsupervisedAD}	& Prediction-based & An LSTM Mixture  Density Network is used to learn the mixture distribution representing the probability density of input data. \\
\hline 
\cite{Inoue2017AnomalyDF} & Prediction-based & An LSTM is used to predict the mean and variance of the input time series. Then the likelihood is used as the anomaly score. \\
\hline 
\cite{Hundman2018DetectingSA} & Prediction-based & The prediction errors are smoothed by EWMA to reduce false positives.\\
\hline 
\cite{deng2021graph} & Prediction-based & Use GAT to predict values of multiple correlated sensors. \\
\hline 
\cite{Lin2020AnomalyDF}	& Prediction-based & A VAE is used to form the local features, and an LSTM is used to estimate the long-term correlation. \\
\hline 
\cite{Liu2020DeepAD} & Prediction-based	& An attention mechanism-based CNN is used to extract features. Then an LSTM is used to calculate the probability of anomaly.\\
\hline 
\cite{dou2019pc} & Prediction-based & The correlation among multivariate time series is presented by ARX models. Then an LSTM is used to predict the abnormal labels of the residual sequence. \\
\hline 
\cite{zhao2020multivariate} & Hybrid & Optimize a reconstruction-based model and a prediction-based model jointly. \\
\hline
\end{tabular}
\end{table*}

The most direct criterion of the reconstruction quality is the reconstruction error, which is the distance between the reconstructed data and the input data\cite{Lee2017AnomalyDI}, as Fig. \ref{fig4.1} shows. Zhang et al. \cite{Zhang2019ADN} used multi-scale reconstruction errors of multivariable time series to diagnose anomalies in complex systems. They constructed multi-scale matrices of system states and built convolutional encoders to embed the temporal dependencies between various sensors into low-dimensional representations. Then a convolutional decoder is used to reconstruct the learned representations. The obtained residual matrix can be applied to anomaly detection, root cause analysis \cite{DNA-analyzer}, and anomaly degree interpretation.

Zhou et al. \cite{zhou2021feature} pointed out that reconstruction can be viewed as a process of projecting a test sample on the training data manifold. They combined the reconstruction error of an autoencoder with the latent representation and reconstruction residual vector to form a new representation for anomaly detection. For a test sample, these three factors correspond to its projection on the training data manifold, its direction to its projection, and its distance to its projection. Therefore, the new representation can characterize how a test sample deviates from the normal pattern, and anomaly detectors based on the new representations can have better generalization performance on unseen data. 

Another criterion of the reconstruction quality is reconstruction probability. Pereira et al. \cite{Pereira2018UnsupervisedAD} utilized a VAE to reconstruct the input data distribution parameters (the mean $\boldsymbol{\mu}_{x}$ and the variance $\boldsymbol{b}_{x}$), and the reconstruction probability is considered the abnormal score. First, they constructed an encoder to obtain posterior parameters ($\boldsymbol{\mu}_{z}$, $\boldsymbol{\Sigma}_{z}$) of an input sequence. Then they sampled $L$ instances from the latent distribution of the VAE and reconstructed their parameters ($\boldsymbol{\mu}_{l_{x}}$, $\boldsymbol{b}_{l_{x}}$). Finally, they computed reconstruction probability, the average log-likelihood of the input data, as the anomaly score. 

In addition to autoencoders, another model commonly used for reconstruction-based anomaly detection is GAN. The generator network of a GAN can generate realistic (fake) time series from the latent space. However, GANs do not directly offer the mapping from the data space to the latent space. Bashar et al. \cite{Bashar2020TAnoGANTS} proposed an iterative search algorithm to find corresponding latent representations of the input time series. Thus, the entire GAN-based framework in \cite{Bashar2020TAnoGANTS} can reconstruct time series and detect anomalies.

To combine the advantages of autoencoders and GANs, Audiber et al. \cite{Audibert2020USADUA} performed adversarial training on two autoencoders. On the one hand, the encoder-decoder structure improves the stability of the adversarial training. On the other hand, the adversarial training allows the model to amplify the reconstruction error of inputs that contain anomalies. Both autoencoders learn to reconstruct a normal sample in the first training stage. Then they compete against each other: $AE_2$ is trained to distinguish the real data from the data generated by $AE_1$, and $AE_1$ is trained to fool $AE_2$. Finally, the anomaly score is a weighted sum of the two autoencoders' reconstruction errors.

\subsubsection{Discrimination-Based Methods}
\begin{figure}[!t]
\centering
\includegraphics[width=3.2in]{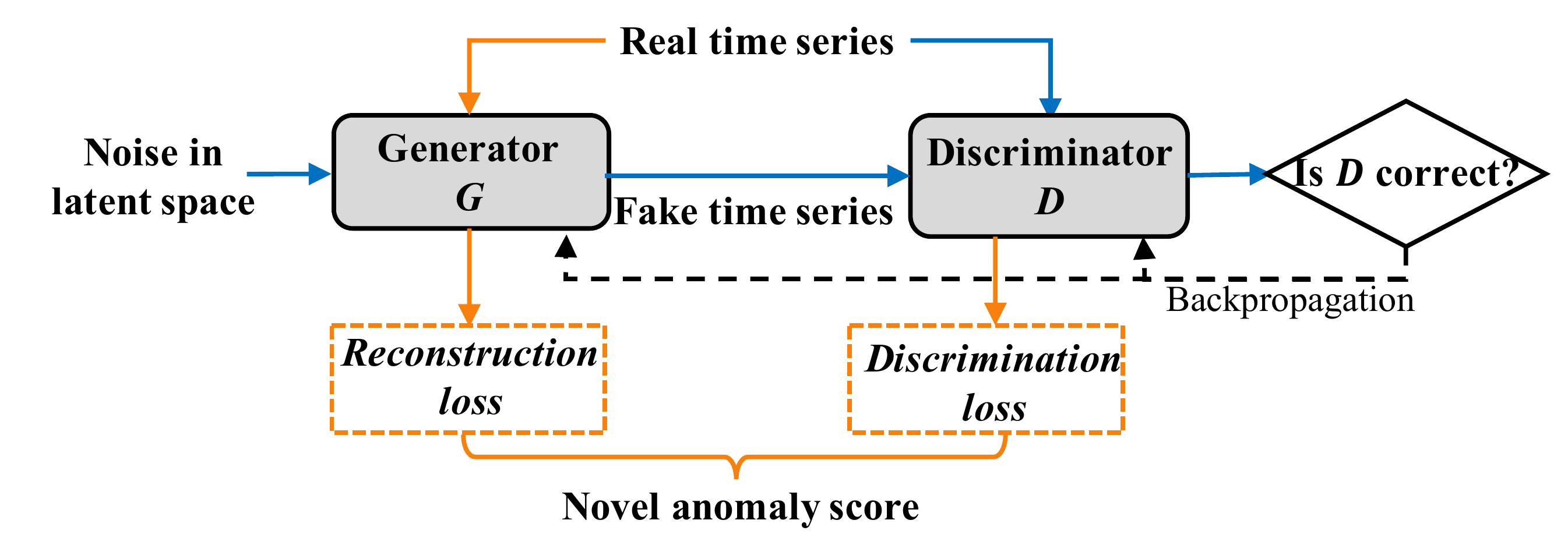}
\caption{A combination of reconstruction loss and discrimination loss.}
\label{fig4.3}
\end{figure}

The core idea of discrimination-based anomaly detection is to establish a discriminator model that directly predicts anomaly scores or anomaly labels of the input time series. For example, a GAN presents a discriminator $D$ that learns to distinguish between fake (abnormal) data and true (normal) data. Li et al. \cite{Li2019MADGANMA} proposed a GAN-based time series anomaly detection framework named MAD-GAN as Fig. \ref{fig4.3}. This framework predicts a novel anomaly score that combines the discrimination results of the discriminator $D$ and the reconstruction residuals of the generator $G$. 

The reconstruction-based and the discrimination-based anomaly detection can efficiently detect subsequence anomalies or sequence anomalies. The following subsection will introduce the prediction-based time series anomaly detection methods, which are more suitable for predicting whether an observation in a time step is abnormal based on contextual information.

\subsubsection{Prediction-Based Methods}
Generally, prediction-based anomaly detection methods work as follows. A prediction model trained on normal samples predicts the values or conditional probabilities of future time steps based on the previous observations. Data points that deviate from the prediction will be deemed as anomalies.

In general, the difference between the incoming value and the predicted value can be used as the anomaly score for each timestamp. The most commonly used model for sequence prediction is RNN, which can learn complex temporal dependencies between previous and current time steps. Lin et al. \cite{Lin2020AnomalyDF} proposed a hybrid model capable of identifying anomalies that span over multiple time scales. Their model uses a VAE to form robust local features in short windows and an LSTM to estimate long-term correlations in the sequence on top of features inferred from the VAE. 

Hundman et al. \cite{Hundman2018DetectingSA} pointed out that the abrupt change of time series is often not perfectly predicted, resulting in sharp spikes in the residual error sequence even when the change is normal. To reduce the false positives of anomaly detection caused by the false alarm spikes, they smoothed the residual error sequence of LSTM by an exponential weighted average (EWMA) algorithm.

For multivariable time series, Deng et al. \cite{deng2021graph} consider sequences from multiple sensors as nodes of a graph and inter-sensor correlations as edges. They forecast future values of each sensor based on a graph attention function over its neighbors. Then they calculated anomaly scores, or prediction errors, for each sensor and each time tick to figure out anomaly states of the whole system. Chen et al. \cite{chen2021learning} also regarded IoT sensors as nodes of a directed graph structure. They used an efficient multi-branch Transformer to make a single-step time series forecasting and return an anomaly score for each testing timestamp. Combining the graph structure with the Transformer enables the model to simultaneously capture inter-variable correlations and temporal dependencies of high-dimensional time series. 

Some anomaly detection methods make predictions on probability distributions of the input data rather than values. These methods define anomalies as observations drawn from a significantly divergent unknown distribution. Inoue et al. \cite{Inoue2017AnomalyDF} constructed an LSTM to predict the mean and variance of the input time series and used the likelihood of the series as the anomaly score. Liu et al. \cite{Liu2020DeepAD} also adopted a similar prediction principle, but they extracted more fine-grained features of the time series by an attention mechanism-based CNN.

Walton et al. \cite{Walton2017UnsupervisedAD} utilized mixture models to predict more complex distributions. They used a $K$ component Gaussian mixture model to approximate the probability density of the digital radio transmissions time series in a dynamic environment. Given a historical sequence, they proposed an LSTM Mixture Density Network (MDN) to estimate the parameters of the Gaussian mixture model, and the likelihood of each test sequence is considered the anomaly score.

Besides, some works directly predict anomaly labels of the test samples. Dou et al. \cite{dou2019pc} focused on collective contextual anomalies (CCA) that break the complex relations among multivariable time series in a complex IT system. They modeled the system with an invariant graph, in which each node represents a univariable time series, and each edge represents a correlation between two nodes. They built autoregressive with exogenous terms (ARX) models to capture each edge's invariance. Then, an LSTM is used to predict the anomaly labels of the residual sequences obtained from the ARX models.

\subsubsection{Hybrid Methods} 
Different anomaly detection methods can complement each other. Zhao et al. \cite{zhao2020multivariate} jointly optimized a reconstruction-based model and a prediction-based model to obtain better time series representations for anomaly detection. Precisely, they summed the loss functions of a single-timestamp prediction and the complete sequence reconstruction. The anomaly score for each timestamp is calculated based on the prediction value and reconstruction probability output by the hybrid model.

\begin{table*}[t]
\renewcommand{\arraystretch}{1.2}
\caption{Summary of DL-based time series clustering methods.}
\label{table4.3}
\centering
\begin{tabular}{c|c|m{0.63\textwidth}}
\hline \hline
Work & Category & Description\\
\hline 
\cite{Richard2020AutoencoderbasedTS} & Deep representation learning-based & Representations learned by a convolutional autoencoder are clustered by K-Medoids.\\
\hline 
\cite{Kalinicheva2020UnsupervisedSI} & Deep representation learning-based & Representations learned by a multi-view 3D convolutional autoencoder are clustered by the hierarchical clustering algorithm (HCA).\\
\hline
\cite{Bhatnagar2017UnsupervisedLO} & Deep representation learning-based & Representations learned by an array of convolutional autoencoders and LSTM autoencoders are clustered by K-Means.\\
\hline
\cite{Ma2019LearningRF} & Joint representation learning-based & The temporal reconstruction and K-means objective are integrated into the seq2seq model to obtain cluster-specific temporal representations. \\
\hline 
\cite{Ienco2020DeepMT} & Joint representation learning-based & A GRU-based autoencoder is used to obtain initial representations of time series, which are stretched towards clustering centroids by K-Means.\\
\hline 
\cite{Tzirakis2019TimeseriesCW} & Joint representation learning-based & A framework is proposed to simultaneously implement video clustering, representational learning, and action segmentation. The clustering process provides supervisory cues for other tasks. \\ 
\hline 
\cite{Trosten2019RecurrentDD} & Label prediction-based & A recurrent network is presented to predict soft clustering labels of the input time series.\\
\hline 
\cite{Tavakoli2020ClusteringTS} & Label prediction-based & The clustering is transformed into a task of label prediction.\\
\hline
\end{tabular}
\end{table*}

\subsection{Models for Clustering}
\label{ClusterTask}
DL-based time series clustering methods can be mainly divided into three categories: deep representation learning-based clustering, joint representation learning-based clustering, and label prediction-based clustering.

\subsubsection{Deep Representation Learning-Based Clustering}
Appropriate time series representation is essential for the efficiency and accuracy of clustering \cite{ratanamahatana2005novel}. If two time series are similar in data space, their representations should also be similar in latent space. Thus, researchers can apply traditional clustering algorithms to the latent representations extracted by DL models. When clustering on low-dimensional representations, the algorithm causes fewer memory requirements and less computational overhead of distance measurements.

Richard \cite{Richard2020AutoencoderbasedTS} learned time series representations with a convolutional autoencoder to accelerate the subsequent clustering. A convolutional autoencoder is trained to reconstruct the input time series. Then a clustering algorithm is applied on top of the learned representations in the latent space of the autoencoder. The clustering algorithm they chose was K-Medoids, which is simple and robust to outliers.

Kalinicheva et al. \cite{Kalinicheva2020UnsupervisedSI} used a hierarchical clustering algorithm (HCA) \cite{Ward1963HierarchicalGT} to cluster the representations of satellite image time series (SITS), which are extracted by a multi-view 3D convolutional autoencoder. The HCA does not demand a researched number of clusters. Besides, Bhatnagar et al. \cite{Bhatnagar2017UnsupervisedLO} used the K-Means algorithm to cluster videos and discover meaningful actions. The video features are extracted in two steps. First, they used an array of convolutional autoencoders to learn the frame-level representations. Then multiple LSTM autoencoders are used to capture temporal information.

\subsubsection{Joint Representation Learning-Based Clustering}
Although the deep representation learning-based clustering methods benefit from the feature extraction capability of DL models, they do not guarantee that the learned representations have a good clustering structure. An effective strategy to solve this problem is jointly learning the representations and clustering. 

\begin{figure}[!t]
\centering
\includegraphics[width=3.2in]{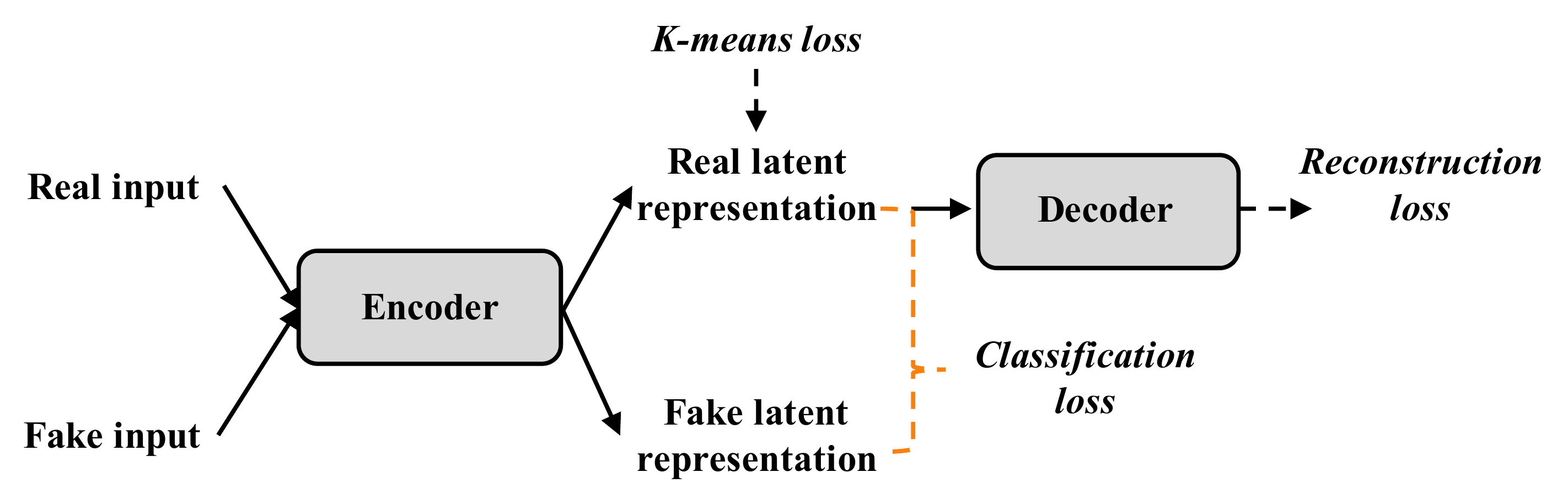}
\caption{A joint representation learning-based clustering framework.}
\label{fig4.4}
\end{figure}

Ma et al. \cite{Ma2019LearningRF} proposed a deep cluster representation (DTCR) framework to obtain time series representations that maintain temporal dynamics, multi-scale features, and good clustering properties. As shown in Fig. \ref{fig4.4}, they used an autoencoder based on bidirectionally expanded RNN to reconstruct the input data. Most importantly, a K-Means objective is integrated into the latent space to guide the representation learning.

Similarly, Ienco and Interdonato \cite{Ienco2020DeepMT} used the K-Means algorithm to stretch the learned representation manifold towards clustering centroids. They obtained the initial representations with a GRU-based autoencoder that consists of two encoders in reverse directions. Then the initial representations are optimized cyclically: 1) during each training epoch, the K-Means algorithm is executed over the current representations; 2) then the distance between the representations and the clustering centroids is added into the loss function. Their hybrid algorithm can handle multivariate time series of variable lengths, but the number of clusters must be specified in advance.

Further, Tzirakis et al. \cite{Tzirakis2019TimeseriesCW} proposed a framework that simultaneously implements video clustering, representation learning, and action segmentation. The framework is divided into three parts: 1) clustering the time series data (this process provides supervisory cues for the proposed framework); 2) learning deep representations in an end-to-end manner with CNNs; 3) identifying the temporal boundaries of the segments. These three parts are optimized iteratively during the training phase.

\subsubsection{Label Prediction-Based Clustering} 
There have been some attempts to transform clustering into a label prediction problem based on DL. Trosten et al. \cite{Trosten2019RecurrentDD} proposed a recurrent network to predict soft clustering labels for the input time series. They first used a two-layer bidirectional GRU to obtain representations for all time series. Then they predicted soft clustering labels for the representations by a fully connected output layer with a softmax activation function. Finally, they defined a divergence-based loss function to discover the underlying clustering structure. This loss function consists of three terms corresponding with three critical characteristics of clustering: 1) cluster separability and compactness; 2) cluster orthogonality in the observation space; 3) closeness of cluster memberships to a simplex corner. This model can handle multivariate time series of variable lengths and does not require distance measurement in the data space.

Tavakoli et al. \cite{Tavakoli2020ClusteringTS} used an autoencoder to predict pseudo labels for data, which are assigned based on the initial clustering results. The neuron in the output layer represents the probabilistic value of the clustering label, and the clustering accuracy can be measured by the mean square error (MSE) of the prediction results.

\emph{Summary:} 
This section systematically reviews DL-based time series anomaly detection and clustering. First, we introduced DL models and techniques commonly used in time series analysis. Then, we discussed DL-based feature extraction. The related works are summarized in Table \ref{table4.1}. Finally, we categorized and reviewed the DL-based approaches for time series anomaly detection and clustering. The involved works are listed in Table \ref{table4.2} and Table \ref{table4.3}.

\section{Applications}
\label{Applications}
With the development of IoT, anomaly detection and clustering have been used widely to explore complex data patterns and provide recommendations to system administrators. This section first introduces the emerging applications of IoT time series anomaly detection and clustering. Then some well-known time series datasets are summarized.

\subsection{Applications of IoT Time Series Analysis}
\label{appForTS}

\subsubsection{Smart Healthcare}

Smart healthcare is a health service system that uses technology such as IoT to access information dynamically and then actively manages and responds to medical ecosystem needs in an intelligent manner \cite{tian2019smart}. The application of smart healthcare contains smart hospitals, assisting diagnosis and treatment, health management, etc. IoT time series analysis has been widely used in smart healthcare. For example, the recent advent of low-cost IoT-based health sensors can produce an enormous amount of time series data for continuous monitoring of various physiological and psychological parameters of a human body \cite{cheng2020leveraging}.

\cite{ray2021iot} proposed an IoT-edge-enabled anomaly detection method to work on the pulse sensor-driven real-life analog health time series data. It can analyze IoT-based sensor-originated health data at the edge devices quickly and automatically. \cite{gupta2021hierarchical} aimed to overcome the drawbacks of centralized anomaly detection models in the Internet of Medical Things (IoMT). They proposed a Federated Learning (FL) based anomaly detection model which utilizes edge cloudlets to run models locally without sharing patients' data. The models analyzed time series from a set of devices, such as motion sensor, smart thermometer, smart oximeter, and smart ECG. \cite{pereira2019learning} proposed an unsupervised anomaly detection method for healthcare time series data to cope with the issue that labels are often difficult to obtain in applications like healthcare.

\cite{bharathi2020energy} presented a particle swarm optimization-based clustering technique for the effective selection of cluster heads among diverse IoT devices. Appropriate cluster head selection can reduce the amount of energy spent on transmitting data from IoT devices to a cloud server. \cite{gupta2021resolving} clustered the sensory data collected through wearable devices to obtain a summarized version of the original data and surmount the data overload and processing latency in real-time remote monitoring.

\subsubsection{The Industrial Internet of Things (IIoT)}
The IIoT is comprised of sensor driven computing, data analytics and intelligent machine applications to provide scalability, efficiency and interoperability which directly promotes automation in critical infrastructure and improve enterprise productivity \cite{hassanzadeh2015towards}. Artificial Intelligence for IT Operations (AIOps) is closely related to the management of IIoT. Data from sensors and equipment in factories is collected and analyzed. For example, the shop floor is tracked and monitored by sensors in real-time, and predictive analytics is used to identify, predict, and prevent. AIOps empowers engineers to efficiently build services in IIoT that are easy to maintain and help to achieve higher service quality and customer satisfaction, increase engineering productivity, and reduce costs \cite{dang2019aiops}.

In IIoT systems, data anomalies inevitably appear due to the scale, computation, and storage complexities. In addition, the networked sensors make IIoT systems more vulnerable to attacks on the control elements, network, or physical environment. Therefore, timely detection of anomalies in sensor readings helps to ensure maximum uptime for machines. \cite{shah2018anomaly} monitored different sensor data of engines, such as fuel usage, engine load, and oil pressure, to detect potential engine failures. Giannoni et al. \cite{giannoni2018anomaly} developed an anomaly detection framework for a wastewater plant where IoT sensors are deployed to manage chemical and particulate concentrations in storage tanks. This framework triggers reactive measures automatically to identify the abnormal state of tanks. Aoudi et al. \cite{aoudi2020scalable} detected subtle structural changes in multivariate monitored signals to prevent cyber-attacks on cyber-physical systems. In addition, \cite{genge2019anomaly} pointed out that it is beneficial to consider the gradual aging of the IIoT's physical dimension when detecting anomalies.

Clustering methods can capture underlying states of industrial time series and identify unexpected events. Sun et al. \cite{sun2019modeling} used a graph-based clustering algorithm to detect botnets in IoT networks based on the assumption that a group of similar nodes in a graph might represent a botnet. Javier et al. \cite{diaz2018clustering} utilized the cluster assignment robustness to detect concept drift in a monitoring system. The concept drifts alert that the monitored process is changing over time, probably due to degradation or other abnormal behaviors.

\subsubsection{Smart Buildings/Smart Cities}
Smart buildings and smart cities use IoT devices to monitor various entities of citizens, devices, buildings, and streets. The collected data is then processed to monitor and manage traffic and transportation systems, air quality, human behavior, cyber-attacks, etc. 

Anomaly detection methods have been applied to detect traffic congestion and incidents. These methods can improve city mobility by regulating vehicular traffic or advising users to modify their path to avoid traffic jams \cite{kong2018lotad}, \cite{d2017detection}. The system in \cite{Biradar2019ChallengesIT} automatically detects anomalous events in traffic videos, which can be applied to traffic rules violation detection and suspicious movements analysis. \cite{el2017monitoring} detected various anomalous road surface conditions, such as potholes, manholes, transverse cracks, decelerating strips, and railroad crossings. 

Environmental pollution management has gradually become a critical issue in smart cities as the population density continues to grow. Jain et al. \cite{jain2016anomaly} analyzed time series of the air pollution monitoring system to detect unhealthy or anomalous locations. The framework in \cite{chen2017adf} can detect potential regional emission sources and identify malfunctioning devices. The government has referred to their analysis results when formulating environmental policies. Clustering methods can also explore valuable information for environmental monitoring. \cite{zhang2016indentifying} identified the major air pollutants of seventy-four Chinese cities based on clustering. \cite{hua2018applied} identified the pattern of air pollution sources using chemometric analysis through hierarchical clustering. \cite{d2015time} clustered air quality time series sampled at different sites to identify similar patterns and reduce redundant information.

The vehicle-to-everything (V2X) \cite{ghosal2020security} is an important application of IoT technologies in the transportation industry and a key component of smart cities. Attacks on vehicles can lead to the leakage of personal information or even traffic accidents \cite{ahmed2018secure}. Therefore, time series anomaly detection and clustering have been widely used in the security of V2X. \cite{qin2021application} proposed an LSTM-based time series anomaly detection framework for the message flows of the in-vehicle CAN network. \cite{sedar2021reinforcement} applied reinforcement learning (RL) approach to detect misbehaving vehicles by exploiting real-time position and speed patterns. \cite{negi2020distributed} proposed a distributed anomaly detection system framework on autonomous vehicle data. \cite{liu2017distributed} presented a stable clustering algorithm for V2X networks to provide traffic information accurately and instantaneously for traffic control.

Another application of IoT time series anomaly detection and clustering is human behavior analysis. Anomaly detection helps detect health problems or risky behaviors of people. The smart assisted-living systems for elderly care proposed in \cite{zhu2015wearable} can effectively detect anomalous behaviors in human daily life. For example, when a monitored individual is found to be on the floor for an extended period, this behavior may suggest a fall or collapse. The framework in \cite{hela2018early} used causal association rules mining to extract anomalous behaviors. For instance, when one is using the phone despite being in the kitchen but the stove is on, there may be a risk of FireElectricity. 

Clustering can discover semantically meaningful actions present in videos, promoting the convenience and safety of smart buildings and smart cities. \cite{Tzirakis2019TimeseriesCW} proposed a graph-degree linkage clustering algorithm for human motion segmentation. The algorithm can analyze whether the monitored object is performing a particular activity. In \cite{Bhatnagar2017UnsupervisedLO}, a robust first-person action clustering approach is proposed to automatically analyze lifelogging videos generated by wearable cameras.

Finally, as more intelligent appliances have been connected to the Internet, these vulnerable IoT devices have become the targets of cyberattacks. \cite{yamauchi2020anomaly, ramapatruni2019anomaly, yamauchi2019anomaly} studied how to identify anomalous activities and attacks in smart buildings.

\subsubsection{Smart Energy}
The real-time monitoring and control of smart grids (SGs) are critical to enhancing power utilities' reliability and operational efficiency. The massive number of time series generated by smart meters (SMs) provides opportunities for better monitoring of power utilities.

Anomalous behaviors in smart grids include transmission line outages, unusual power consumption, momentary and sustained outages. Anomaly detection methods for SGs can be divided into four main categories \cite{moghaddass2017hierarchical}: 1) consumption analysis; 2) malicious and security attacks detection; 3) fault location; and 4) outage detection. In \cite{moghaddass2017hierarchical}, operators monitored power usage to enhance the situational awareness of utility operators so that they could identify faults in the local distribution network in real-time before customers' feedback. 

Passerini et al. \cite{passerini2019smart} proposed a framework that enables the autonomous detection and location of network anomalies in distribution grids. The framework mainly includes two algorithms. The first algorithm is used to detect and track the evolution of faults over time. The second algorithm uses the knowledge of the network topology to localize the detected anomaly by analyzing the sensed trace in the time domain. As SMs are likely to be exposed to multiple cyber-attacks, \cite{yip2018anomaly} evaluated consumers' energy utilization behavior to identify potential energy frauds and faulty meters.

The clustering analysis also has great potential in energy efficiency programs. \cite{lavin2015clustering} clustered accounts based on their usage profiles to find accounts with similar energy usage tendencies. \cite{Richard2020AutoencoderbasedTS} clustered electricity consumption time series of different clients to distinguish different types of users (such as residential clients, SMEs, and secondary homes). This process is conducive to the refined management of the energy grids.

\subsection{IoT Time Series Datasets}
Various IoT time series datasets have been proposed in recent years. This subsection introduces some IoT time series datasets commonly used in anomaly detection and clustering, as summarized in Table \ref{table5.1}.

\begin{table*}[t!]
\renewcommand{\arraystretch}{1.2} 
\caption{Summary of IoT time series datasets.}
\label{table5.1}
\centering
\begin{tabular}{m{0.15\textwidth}<{\centering}|c|c|m{0.65\textwidth}}
\hline \hline
Name & Year & Ref. & Description \\
\hline
SWaT & 2015 & \cite{Goh2016ADT} & A water treatment time series datasets with simulated attack scenarios. It monitors 51 sensors for 11 consecutive days. \\
\hline
WADI & 2016 & \cite{WADIdataSet} & A water treatment time series datasets with simulated attack scenarios. 
It monitors 103 sensors for 16 consecutive days.\\
\hline
WESAD & 2018  & red{\cite{schmidt2018introducing}} & A multivariate time series dataset of wearable stress and affect detection. It can be used in smart healthcare.\\
\hline
BoT-IoT & 2019 & \cite{koroniotis2019towards} & An IoT botnet dataset that incorporates legitimate and simulated IoT network traffic, along with various types of attacks. \\
\hline
IoTID20 & 2020 & \cite{ullah2020scheme} & An IoT anomalous activity detection dataset generated through home-connected smart devices. It includes eight attack types. \\
\hline
MQTTset & 2020 & \cite{vaccari2020mqttset } & An IoT dataset focused on MQTT communications and the associated IoT threats. \\
\hline
MedBIoT & 2020 & \cite{guerra2020medbiot}	& An IoT botnet detection dataset collected from eighty-three real and emulated IoT devices. \\
\hline
IoT-23 & 2020 & \cite{parmisano2020labeled } & A labeled dataset with malicious and benign IoT network traffic over 2018 to 2019 from twenty-three different scenarios. \\
\hline
MQTT-IoT-IDS2020 & 2021 & \cite{hindy2021machine} & A dataset collected from MQTT-IoT sensors. The network consists of twelve MQTT sensors and contains four types of attacks. \\
\hline
TON\_IoT & 2021 & \cite{moustafa2021new} & Datasets that include heterogeneous telemetry data of IoT/IIoT services, as well as the operating systems logs and network traffic of IoT network. \\
\hline
X-IIoTID & 2021 & \cite{al2021x } & An IIoT intrusion dataset with the behaviors of new IIoT connectivity protocols, activities of recent devices, diverse attack types and scenarios, and various attack protocols. \\
\hline
IoTHealth & 2021 & \cite{IoTHealth} & A synthetic IoT time series dataset for smart healthcare. Five human physiological parameters are considered. It can be used for multivarite or univarite anomaly detection. \\
\hline
CIC IoT dataset 2022 & 2022 & \cite{dadkhah2022towards } & A dataset for IoT identification/profiling and intrusion detection. The network contains sixty IoT devices. Different stages, scenarios and attacks are analyzed. \\
\hline
Edge-IIoT & 2022 & \cite{ferrag2022edge } & A comprehensive realistic cyber security dataset of IoT and IIoT applications for centralized and federated learning. Five threats are included. \\

\hline
\end{tabular}
\end{table*}

Secure Water Treatment Dataset (SWaT) \cite{Goh2016ADT} and Water Distribution Dataset (WADI) \cite{WADIdataSet} are two commonly used IoT time series datasets for anomaly detection. SWaT is a testbed for cyber-security research built at the Singapore University of Technology. The dataset contains 51 variables (sensor readings and actuator status) for 11 consecutive days: 7 days collected under normal operations and 4 days collected with attack scenarios, during which 36 simulated attacks were carried out. The WADI dataset is collected from the WADI testbed by measuring 103 variables for 16 days. The first 14 days are under normal operations, and the last 2 days are under attack scenarios.

The Wearable Stress and Affect Detection (WESAD) dataset \cite{ schmidt2018introducing } records physiological and motion data of fifteen subjects measured through two IoMT devices, namely RespiBAN and Empatica E4, for two hours. The data includes the following sensor modalities: blood volume pulse, electrocardiogram, electrodermal activity, electromyogram, respiration, body temperature, and three-axis acceleration.

The BoT-IoT dataset \cite{koroniotis2019towards} is an IoT botnet dataset collected from a simulated IoT environment. The testbed applied five IoT scenarios: a weather station, a smart fridge, motion-activated lights, a remotely activated garage door, and a smart thermostat. The BoT-IoT dataset contains over seventy-two million incorporated legitimate and simulated IoT network traffic, along with various attacks, such as DDoS, DoS, service scan, keylogging, and data exfiltration.

The IoTID20 dataset \cite{ullah2020scheme} is an IoT botnet dataset generated from a testbed of a smart home environment which consists of smart home device SKT NGU and EZVIZ Wi-Fi camera. Other devices connected to the smart home router include laptops, tablets, and smartphones. The SKT NGU and EZVIZ Wi-Fi camera are IoT victim devices, and all other devices in the testbed are the attacking devices. Eight types of attacks were conducted, such as Syn flooding, host brute force, and ARP spoofing. The final version of the IoTID20 dataset consists of eighty-three network features and three label features.

The MQTTset dataset \cite{ vaccari2020mqttset} is an IoT dataset focused on MQTT communications and the associated IoT threats. It is composed of IoT devices of different natures to simulate a smart home/office/building environment. Eight sensors located into two separated rooms record temperature, humidity, motion, CO-Gas, door opening/closure, fan status, smoke, and light. MQTTset includes both legitimate and malicious traffic over a week. Each sensor is configured to trigger communication at a specific time to simulate a real behavior of a home automation. The malicious traffic was generated by launching attacks against the MQTT broker.

The MedBIoT dataset \cite{ guerra2020medbiot} is an IoT botnet detection dataset collected from real and emulated IoT devices in a medium-sized network (i.e., eighty-three devices). Three actual botnet malware, Mirai, BashLite, and Torii, were deployed in the network, and the dataset is focused on the early stages of botnet deployment (spreading and C\&C communication). MedBIoT is split according to the traffic source (i.e., normal or malware traffic) allowing to easily label the data and extract features from the raw pcap files.

The IoT-23 dataset \cite{ parmisano2020labeled } consists of twenty-three captures (called scenarios) of different IoT network traffic ranging from 2018 to 2019. Both malicious network traffic and benign IoT traffic are included. Specifically, the malicious traffic is obtained from twenty malware captures executed in a Raspberry Pi, and the benign network traffic was obtained from three different real IoT devices: a smart LED lamp, a home intelligent personal assistant, and a smart doorlock.

The MQTT-IoT-IDS2020 dataset \cite{hindy2021machine} is an Intrusion Detection Systems (IDS) dataset based on Message Queuing Telemetry Transport (MQTT) communication protocol. The network consists of twelve MQTT sensors, a broker, a machine to simulate camera feed, and an attacker. The dataset consists of five recorded scenarios: normal operation and four attack scenarios. During normal operation, all sensors send randomized messages with different lengths to simulate different usage scenarios. The attacker performs four types of attacks: aggressive scan, UDP scan, Sparta SSH brute-force, and MQTT brute-force attack.

The TON-IoT datasets \cite{moustafa2021new} include heterogeneous telemetry data of IoT/IIoT services, as well as the operating systems logs and network traffic of IoT network, which were collected from a realistic representation of a medium-scale network. In the testbed of TON\_IoT, two smartphones and a smart TV were logged in network traffic, and seven IoT and IIoT sensors (e.g., weather, temperature, and Modbus sensors) were used to capture their telemetry data. The network dataset of Ton\_IoT includes nine types of attacks, such as scanning, DoS, DDoS, and ransomware.

The X-IIoTID dataset \cite{ al2021x } is the first-of-its-kind IIoT intrusion dataset that includes the behaviors of new IIoT connectivity protocols, activities of recent devices, diverse attack types and scenarios, and various attack protocols. The author distilled a generic attack life-cycle framework for IIoT attacks and generated different attacks in each stage. X-IIoTID contains 421,417 normal records, 399,417 malicious records, and fifty-nine features collected from network traffic, device resources, and device/alert logs.

The IoTHealth dataset \cite{IoTHealth} can be used for performing multivariate or column-wise univariate anomaly detections. The IoT-based synthetic data considers five human physiological parameters such as skin conductance (C) in micro Siemens, body temperature (T) in Fahrenheit, blood pressure low (BL), blood pressure high (BH), and root mean square of successive difference (RMSSD) of heart rate variability (HRV) in milliseconds. Each of the variable columns include anomalies.

The CIC IoT dataset 2022 \cite{ dadkhah2022towards } is a state-of-the-art dataset for intelligent identification and intrusion detection of sixty different IoT devices with different protocols such as IEEE 802.11, Zigbee-based, and Z-Wave. The data contains different stages of each device and different scenarios of the simulated network activity of a smart home. Besides, two different attacks were performed to capture the attack network traffic.

The Edge-IIoT dataset \cite{ ferrag2022edge } is a comprehensive, realistic cyber security dataset of IoT and IIoT applications for centralized and federated learning. The IoT data were generated from more than ten types of IoT devices, such as low-cost digital sensors for sensing temperature and humidity, ultrasonic sensors, and water level detection sensors. The data features were extracted from different sources, including alerts, system resources, logs, and network traffic. Besides, five threats related to IoT and IIoT connectivity protocols were analyzed, including DoS/DDoS attacks, information gathering, man in the middle attacks, injection attacks, and malware attacks.

\section{Future Challenges and Directions}
\label{Directions}
In this section, we discuss the challenges faced by, and future research opportunities in applying DL in IoT time series analysis.

\subsubsection{Insufficient Labels}
Since the scale and complexity of IoT systems continue to grow, labeling large amounts of time series requires human resources that most organizations cannot afford. Therefore, it is unpractical to train DL models in a supervised manner. Developing accurate and robust unsupervised models is a promising direction. In addition, data augmentation, transfer learning, and meta-learning \cite{hospedales2020meta} are also effective ways to deal with the insufficiency or imbalance of training data.

\subsubsection{Real-Time Performance}
Time series analysis is usually applied in short-term decisions of scenarios such as IIoT, smart cities, and smart energy. Most traditional DL models have limited applications in these time-sensitive tasks because of their high computational complexity. For example, suppose an anomaly detector in an IIoT system takes a long time to process an observation. In that case, the system may have suffered a severe fault before the operator received the alarm. To improve the real-time performance of DL models, researchers can develop lightweight models and algorithms to reduce computational complexity. Besides, the compression and acceleration methods of DL models \cite{choudhary2020comprehensive} are also worthy of attention.

\subsubsection{Interpretability} Generally, interpretability refers to the extent of human's ability to understand and reason a model \cite{fan2021interpretability}. Although DL methods have achieved great success in many fields, their extremely complicated structures make it difficult to understand the innerworkings. Therefore, the lack of interpretability has become one of the primary obstacles of DL methods, especially deep neural networks (DNNs), in their wide acceptance in mission-critical applications such as IIoT. At present, the interpretability of neural networks can be classified into \emph{Post-hoc} interpretability analysis and \emph{ad-hoc} interpretable modeling \cite{fan2021interpretability}. In the future, combining DL with human knowledge or neuroscience may promote interpretable research on DL.

\subsubsection{Distributed System}
A large amount of time series data is generated in a distributed way, such as the observation values from different IoT devices in a factory. However, many DL models require high computational power that significantly outweighs the capacity of resource- and energy-constrained IoT devices, making it difficult to run DL models directly on IoT devices. One common way to solve this problem is to transmit data from IoT devices to a data center, which brings a high additional transmission delay. Edge intelligence is a new technology to achieve low-latency data processing for IoT devices. It moves cloud computing capabilities in data centers closer to the distributed intelligent devices \cite{zhou2019edge}. 

\subsubsection{Irregular Time Series}
Various sequence modeling techniques have been applied in time series analysis, and researchers constantly develop novel algorithms. However, DL models for time series in unconventional forms, such as irregular time series, are relatively unexplored. Irregular data and the resulting missing values severely compromise traditional DL methods. Jiao et al. \cite{jiao2020timeautoml} utilized Automatic Machine Learning (AutoML) to search the optimal neural network structures for irregular time series. In future research, more DL models adapted to irregular time series should be developed and gated recurrent neural networks have been found to be a promising direction \cite{weerakody2021review}.

\subsubsection{Privacy}
IoT devices typically record sensitive information in factories, businesses, homes, and other environments. Privacy breaches may result in severe economic losses and physical threats. Therefore, privacy-protected DL methods are urgently needed in practice. For example, Yang et al.\cite{yang2020privacy} assumed that the controller company has no direct access to users’ consumption requirements during the energy scheduling, which preserves the consumption data privacy. The federated learning (FL) technology \cite{konevcny2016federated} is a promising privacy-preserving machine learning paradigm, in which multiple clients (such as mobile devices) train models collaboratively under a central server (such as a service provider). Other privacy-preserving mechanisms such as differential privacy are also effective in practice \cite{mireshghallah2020privacy}.

\section{Conclusion}
\label{Conclusion}
In the 5G era, IoT networks and mobile applications generate ever-increasing amount of time series data. In the past decade, DL has shown great potential in automatically extracting data features and tackling complex problems in an end-to-end fashion. In this article, we systematically review DL-based unsupervised anomaly detection and clustering under a unified framework. We hope our work can offer insights into the structures and capabilities of DL models and promote the applications of DL in IoT time series analysis.

\bibliographystyle{IEEEtran}
\bibliography{myreference}

%

\begin{IEEEbiography}[{\includegraphics[width=1in,height=1.25in,clip,keepaspectratio]{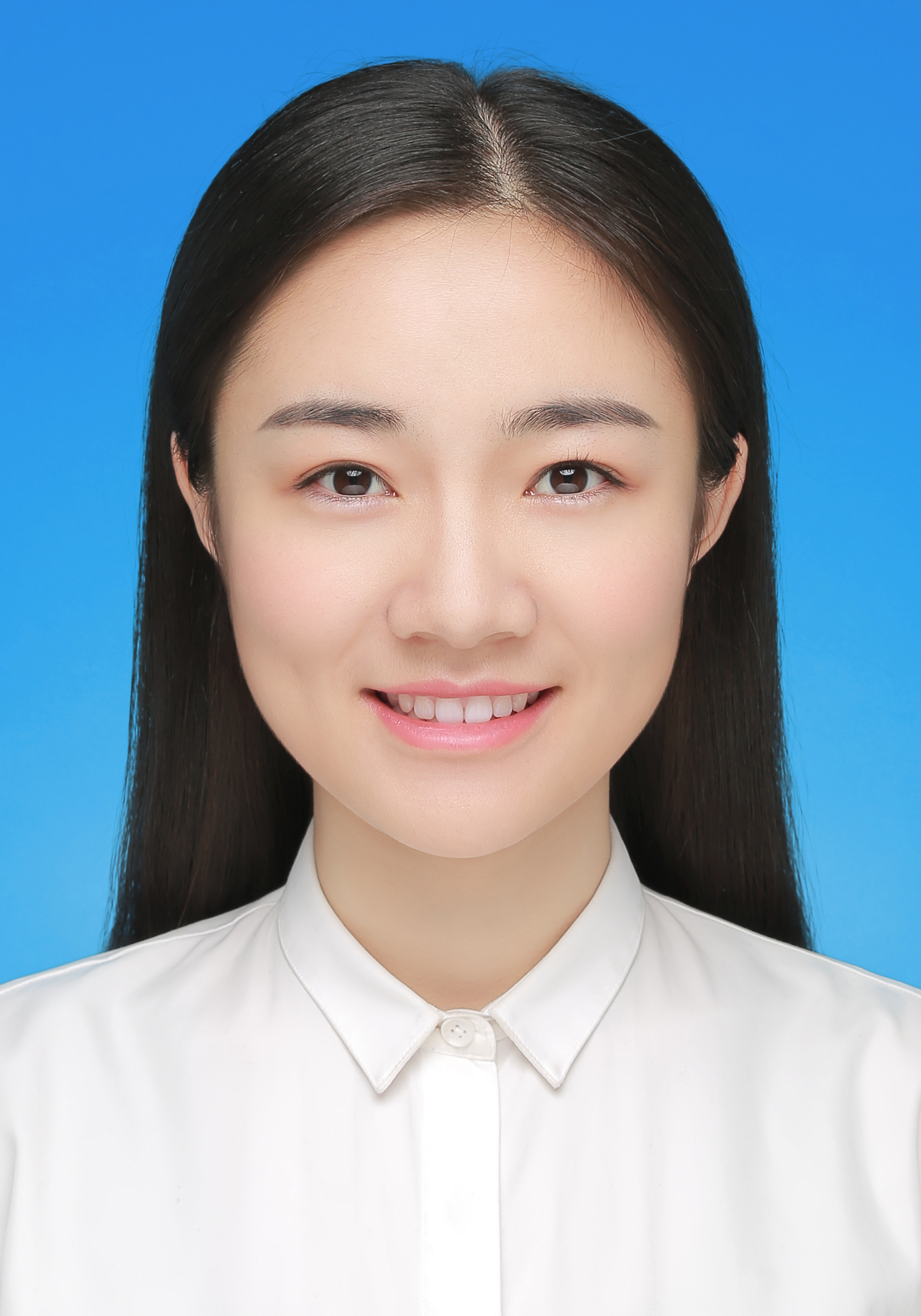}}]{Ya Liu} was born in Shanxi, China in 1997. She received the B.S degree from Beijing Jiaotong University, Beijing, China,
in 2019. She is currently pursuing the Ph.D. degree in computer science from the Department of Computer Science, Tongji University, Shanghai, China.

Her current research interests include anomaly detection, network security, and distributed learning.
\end{IEEEbiography}

\vspace{11pt}

\begin{IEEEbiography}[{\includegraphics[width=1in,height=1.25in,clip,keepaspectratio]{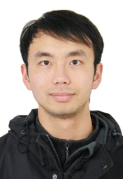}}]{Yingjie Zhou} (M’14) received his Ph.D. degree in the School of Communication and Information Engineering from University of Electronic Science and Technology of China (UESTC), China, in 2013. He is currently an associate professor in the College of Computer Science at Sichuan University (SCU), China. He was a visiting scholar in the Department of Electrical Engineering at Columbia University, New York. His current research interests include network management, behavioral data analysis, and resource allocation. He has served as Program Vice-Chair of IEEE HPCC, Local Arrangement Chair of IEEE BMSB, and TPC member for many major IEEE conferences, such as GLOBLECOM, ICC, ITSC, MSN and VTC. He received the Best Paper Awards at IEEE HPCC and IEEE MMSP in 2022.
\end{IEEEbiography}

\vspace{11pt}

\begin{IEEEbiography}[{\includegraphics[width=1in,height=1.25in,clip,keepaspectratio]{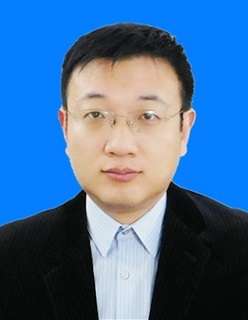}}]{Kai Yang} (SM'18) received the B.Eng. degree from Southeast University, Nanjing, China, the M.S. degree from the National University of Singapore, Singapore, and the Ph.D. degree from Columbia University, New York, NY, USA.

He is a Distinguished Professor with Tongji University, Shanghai, China. He was a Technical Staff Member with Bell Laboratories, Murray Hill, NJ, USA. He has also been an Adjunct Faculty Member with Columbia University since 2011. He holds over 20 patents and has been published extensively in leading IEEE journals and conferences. His current research interests include big data analytics, machine learning, wireless communications, and signal processing.

\end{IEEEbiography}

\vspace{11pt}

\begin{IEEEbiography}[{\includegraphics[width=1in,height=1.25in,clip,keepaspectratio]{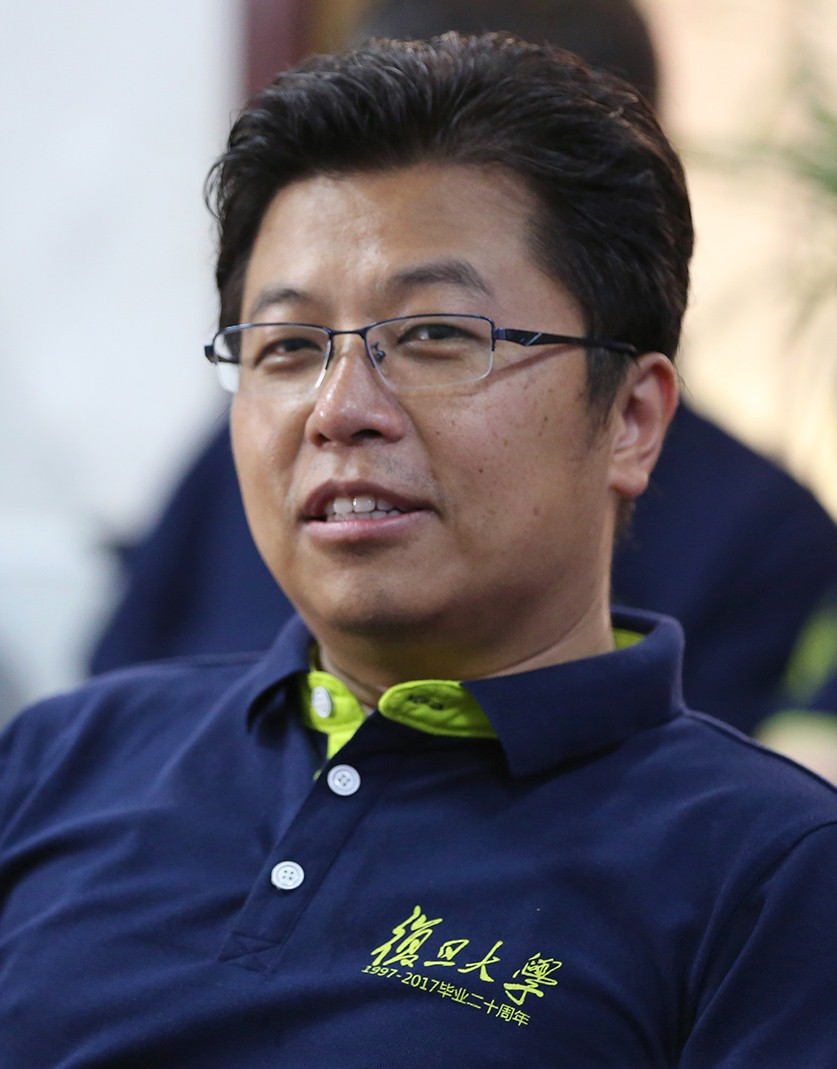}}]{Xin Wang} (SM'09-F'23) received the B.Sc. and M.Sc.degrees from Fudan University, Shanghai, China, in 1997 and 2000, respectively, and the Ph.D. degree from Auburn University, Auburn, AL, USA, in 2004, all in electrical engineering.

From September 2004 to August 2006, he was a Postdoctoral Research Associate with the Department of Electrical and Computer Engineering, University of Minnesota, Minneapolis. In August 2006, he joined the Department of Electrical Engineering, Florida Atlantic University, Boca Raton, FL, USA, as an Assistant Professor, then was promoted to a tenured Associate Professor in 2010. He is currently a Distinguished Professor and the Chair of the Department of Communication Science and Engineering, Fudan University, China. His research interests include stochastic network optimization, energy-efficient communications, cross-layer design, and signal processing for communications. He is a Senior Area Editor for the IEEE Transactions on Signal Processing and an Editor for the IEEE Transactions on Wireless Communications, and in the past served as an Associate Editor for the IEEE Transactions on Signal Processing, as an Editor for the IEEE Transactions on Vehicular Technology, and as an Associate Editor for the IEEE Signal Processing Letters. He is a member of the Signal Processing for Communications and Networking Technical Committee of IEEE Signal Processing Society, and a Distinguished Speaker of the IEEE Vehicular Technology Society.
\end{IEEEbiography}

\vfill





\end{document}